\documentclass[a4paper,10pt,twoside]{article}
\usepackage[top=3cm, bottom=3cm, inner=3cm, outer=3cm, includehead]{geometry}
\usepackage{fancyhdr}
\usepackage{xurl}
\usepackage{graphicx}
\usepackage{alltt}
\usepackage{amsmath}
\usepackage[hidelinks, pdftex]{hyperref}
\usepackage[T1]{fontenc}
\usepackage[utf8]{inputenc}
\usepackage{lmodern}
\usepackage{csquotes}

\usepackage[figuresleft]{rotating}
\usepackage{newtxtext,newtxmath}
\usepackage{graphicx}
\usepackage{float}

\usepackage{amsmath}
\usepackage{amssymb}
\usepackage[export]{adjustbox}
\usepackage{textgreek}
\usepackage[per-mode=symbol]{siunitx}
\usepackage{graphicx} 
\usepackage{array} 
\usepackage{subcaption}
\usepackage{longtable}
\usepackage{booktabs}
\usepackage{enumitem} 
\usepackage{nomencl}
\usepackage[normalem]{ulem}
\usepackage{textcomp}

\usepackage[per-mode=symbol]{siunitx}
\usepackage[backend=biber, sorting=none]{biblatex} 
\usepackage[english]{babel}

\begin{document}
\fancyhead[LE]{\thepage\ \ \ \ Plastropoulos, Avdelidis, and Zolotas}
\fancyhead[RO]{Techno-Economic Analysis\ \ \ \ \thepage}
\begin{center}
\LARGE
\textbf{Techno-Economic analysis for Smart Hangar inspection operations through Sensing and Localisation at scale}\\[12pt]
\normalsize

\textbf{Angelos Plastropoulos\textsuperscript{1,*}, Nicolas P. Avdelidis\textsuperscript{2}, and Argyrios Zolotas\textsuperscript{1}}\\[4pt]
\end{center}

\begingroup
\renewcommand\thefootnote{\fnsymbol{footnote}}
\footnotetext[1]{Corresponding author: a.plastropoulos@cranfield.ac.uk}
\endgroup
\footnotetext[1]{Centre for Assured And Connected Autonomy, Faculty of Engineering and Applied Sciences, Cranfield University, Cranfield, MK43 0AL, United Kingdom}
\footnotetext[2]{Department of Aeronautics and Astronautics, School of Engineering, University of Southampton, Southampton, SO16 7QF, United Kingdom}

\begin{abstract}
\normalsize
The accuracy, resilience, and affordability of localisation are fundamental to autonomous robotic inspection within aircraft maintenance and overhaul (MRO) hangars. Hangars typically feature tall ceilings and are often made of materials such as metal. Due to its nature, it is considered a GPS-denied environment, with extensive multipath effects and stringent operational constraints that collectively create a uniquely challenging environment. This persistent gap highlights the need for domain-specific comparative studies, including rigorous cost, accuracy, and integration assessments, to inform a reliable and scalable deployment of a localisation system in the Smart Hangar. This paper presents the first techno-economic roadmap that benchmarks motion capture (MoCap), ultra-wideband (UWB), and a ceiling-mounted camera network across three operational scenarios: robot localisation, asset tracking, and surface defect detection within a 40 × 50 m hangar bay. A dual-layer optimisation for camera selection and positioning framework is introduced, which couples market-based camera-lens selection with an optimisation solver, producing camera layouts that minimise hardware while meeting accuracy targets. The roadmap equips MRO planners with an actionable method to balance accuracy, coverage, and budget, demonstrating that an optimised vision architecture has the potential to unlock robust and cost-effective sensing for next-generation Smart Hangars.\vskip 2mm
\textbf{Keywords:} aircraft maintenance, motion capture systems, ultra-wideband systems, localisation,
smart hangar, monitoring, defect detection, MRO, robotic inspection, hangar of the
future, smart infrastructure.
\end{abstract}

\section{Introduction}
Accurate localisation is fundamental to enabling autonomous robotic inspection in MRO hangars, where metallic structures, extensive multipath effects, and strict operational constraints define a unique and challenging environment \cite{Plastropoulos2025AHangar}. Existing localisation technologies, including infrared MoCap, UWB real-time location systems, and camera-based or Simultaneous Localisation and Mapping (SLAM) approaches, offer different trade-offs in terms of achievable accuracy, infrastructure complexity, cost, and robustness to occlusion and interference \cite{Masiero2019APositioning}. Sensor fusion frameworks that combine vision, inertial, and UWB data can improve robustness and deliver cost–accuracy trade-offs, as demonstrated in large-scale warehouse deployments. However, their performance and economic viability remain highly dependent on environment-specific factors \cite{Gerwen2022IndoorFusion}. In addition, this study also treats localisation as the backbone for real-time asset monitoring, allowing ground support equipment, tooling, and spares to be tracked seamlessly in the bay. On the opposite scale, artefact localisation is addressed, with an exploration of how sensing can facilitate the identification of surface defects and other critical features on the airframe. The framing of these macroscopic and microscopic needs together sets the stage for the optimisation framework, comparative experiments, and cost analyses developed in the remainder of the paper.

Despite technical advances, a notable lack of comprehensive, real-world techno-economic analyses remains, specifically focused on aircraft hangar deployments.  The available literature provides only partial benchmarking or component-level comparisons for individual or hybrid localisation modalities \cite{Park2023FusionRTLS, Pugliese2021LiDAR-AidedFuselages}, with little empirical evidence on their robustness to the full spectrum of hangar-specific challenges such as dynamic occlusion, specular reflections, and integration with existing maintenance workflows. To date, no studies have presented a holistic side-by-side evaluation of MoCap, UWB, and vision-based solutions in an operational metallic aircraft hangar context. This persistent gap underscores an urgent need for domain-specific comparative studies, including rigorous assessments of cost, accuracy, and integration, to inform the reliable and scalable deployment of robotic inspection systems in aviation environments. The economic impact of daily maintenance practices has begun to be quantified by recent hangar-focused studies; for instance, significant rework costs in narrow-body bays can be avoided by improving technicians' awareness of composite repair, as demonstrated by Jong et al. \cite{Jong2024HandlingPersonnel}. Across the timeline, Moenck et al. \cite{Moenck2025IndustryOpportunities} outline how the forthcoming Industry 5.0 automation could reshape the trade-offs of labour hours and logistics on large MRO campuses. At the same time, the classic aerodynamic analysis of the enclosed engine test hangars by Wallis and Ruglen still provides a valuable historical baseline for energy throughput economics \cite{Wallis1966OnFacilities}.

In summary, this study offers five significant contributions: (i) it presents the inaugural techno-economic roadmap that evaluates MoCap, UWB, and ceiling-camera vision in parallel for full-scale aircraft hangars; (ii) it introduces a dual-layer optimisation framework that combines market-driven camera-lens selection with a Mixed-Integer Linear Programming-based set-cover placement, resulting in the minimal number of cameras needed; (iii) it supplies quantified design-to-cost case studies converting three typical MRO tasks into specific bills of materials and cost estimates; (iv) determines the optimal balance for defect-detection accuracy, illustrating how ceiling cameras and drone close-ups converge at various defect sizes; and (v) provides the first cost/accuracy comparison between camera localisation and commercial UWB/MoCap systems for a conventional 40 × 50 m bay. Collectively, these contributions deliver an actionable and comprehensive methodology for MRO decision-makers to select, size, and cost localisation and inspection systems within large hangars.

Inspired by the aviation industry's shift towards Industry 5.0, which sees mobile robots and AI-based decision support systems taking on routine maintenance duties, the hangar should transition from a passive shelter to a dynamic sensing platform. The end goal is a ceiling infrastructure dense enough to localise robots, track assets, and even surface defects in real-time but lean enough to be economically retrofitted into legacy bays. Against this backdrop, the remainder of the paper is organised as follows. Section 2 reviews the state-of-the-art in MoCap, UWB and ceiling-camera vision, clarifying their respective accuracy, cost, and integration trade-offs. Section 3 introduces a dual-layer optimisation framework that first selects a market-ready camera–lens pair and then solves a set-cover problem to minimise hardware while meeting resolution targets. Section 4 translates those algorithms into three design-to-cost frameworks: robot localisation, asset tracking, and defect detection, each with a detailed bill of materials for practitioner guidance. It also presents MoCap and UWB implementation options for benchmarking.

\section{Digital Sensing Options for the Smart Hangar}
The hangar is a space where multiple maintenance tasks from various technical teams are performed simultaneously. Effective coordination, spatial awareness, and asset management are crucial to increase productivity and minimise mistakes. In addition, precise localisation is crucial for evaluating applications involving robotics. Therefore, it is no surprise that when Airbus first introduced the concept of the "Hangar of the Future", the localisation system was part of the suggested enabling technologies \cite{Plastropoulos2024TheAviation}. For this concept, the UWB system was chosen. The choice was reasonable since it provides a balance between cost and accuracy. Ubisense has shared two related experiments. In the first, the system provides localisation data for a drone that performs aircraft inspection \cite{UbisenseAviationGuide}. In the second study \cite{UbisenseUbisenseSolution}, the focus was on asset management and digital twins. An alternative method, which is not widely adopted in hangars, involves utilising a MoCap system. Although this provides greater accuracy, it is more expensive and requires a line of sight with a minimum of two cameras. Yet another technique, still in the research stage but showing significant potential, is to explore the usage of a camera-based system enhanced with deep learning models capable of executing various tasks. The subsequent section outlines each approach by highlighting its fundamental functions, characteristics, and constraints, along with its standard architecture. For the camera-based system, three candidate architectures are illustrated on the basis of specific target scenarios.

\subsection{Optical Motion-Capture}
MoCap systems are technologies that estimate the 3D positions and orientations (6-DOF poses) of tracked objects using camera-based tracking of reflective or active markers. These systems operate by triangulating marker positions from multiple synchronised camera views and are widely used in fields such as robotics, biomechanics, and animation. Commercial systems such as Qualisys, Vicon, and OptiTrack use high-speed infrared cameras configured around a workspace to track marker-equipped objects with submillimeter precision. For robotic localisation tasks, especially in indoor settings, MoCap systems provide highly accurate pose data that can be integrated into robotic frameworks, such as those based on the Robot Operating System (ROS), enabling real-time control, mapping, or trajectory following.

The MoCap systems use multiple infrared cameras and reflective markers to determine 3D positions in real-time. Reflective markers are small spheres (often coated with retro-reflective material) placed on the object of interest. The reflective coating ensures that the markers appear as bright spots in the camera images whenever IR light is reflected from them. For robot localisation, typically, multiple markers are affixed in a fixed configuration on the robot, forming a rigid body. A typical MoCap setup (Figure \ref{fig:topologies}a) involves several hardware components, such as infrared cameras with built-in IR LED strobe rings, synchronisation modules, data acquisition hardware, and a central processing computer. The architecture ensures that all cameras capture movement simultaneously and feed data to a hub that computes the position and orientation of the robot at high update rates. For accurate 3D reconstruction, all cameras must capture frames in sync. A synchronisation unit ensures that each camera exposure is synchronised in time. This hardware unit maintains a consistent frame rate across all cameras, ensuring that the position of a fast-moving object is recorded at the same instant from all viewpoints. Optical MoCap cameras typically connect to a data acquisition network that handles both data transfer and power supply. This is often achieved via Power over Ethernet (PoE). A PoE network switch serves as the central hub to which all cameras connect via Ethernet cables. This switch provides DC power to the cameras via the same cable that carries data, simplifying installation (no separate power cords are required for each camera). Finally, at the centre of the system is the processing hub, typically a powerful host computer that runs MoCap software. This unit is where all camera data are converged for real-time processing and storage.

\begin{figure}[h!]
    \begin{tabular}{ c p{0.35\textwidth} }
      \adjincludegraphics[width=0.6\textwidth, valign=c]{images/mocap.PNG} &
      (a) System architecture of an indoor optical MoCap setup for robot localisation. Multiple IR cameras observe reflective markers on the robot, all synchronised and connected via a network to a central processing computer that computes the robot’s 3D position and orientation. \\
      \\
      \adjincludegraphics[width=0.6\textwidth, valign=c]{images/uwb.PNG} &
      (b) System architecture of a UWB system for robot localisation. Anchors are fixed in the hangar and report each tag's position (x,y,z) in the operational space. \\
      \\
      \adjincludegraphics[width=0.6\textwidth, valign=c]{images/camera.PNG} &
      (c) System architecture of a camera-based system. Each camera is connected through a PoE Ethernet network switch to the central processing unit. In this central node, multiple deep learning machine vision models will be executed in order to estimate the required data based on the target scenario. 
      \\
    \end{tabular}
    \centering
    \caption{The system architectures of the three suggested architectures.}
    \label{fig:topologies}
\end{figure}

In robotic applications, particularly in large indoor environments such as industrial inspection areas or aircraft hangars, the deployment of MoCap systems faces a number of challenges. These include maintaining camera visibility across wide areas, synchronising data from distributed cameras, mitigating occlusion or misalignment of markers, and ensuring system accuracy despite environmental interference such as lighting or reflective surfaces. 

Among the studies reviewed, commercial systems have been extensively benchmarked under controlled conditions, showing static accuracy of approximately 0.15 mm and dynamic errors of less than 2 mm \cite{Merriaux2017APerformance}, although these results stem from experiments not explicitly conducted in large-scale industrial settings. Studies such as \cite{Rahimian2017OptimalSystems} have proposed optimised camera placement algorithms to improve tracking performance under challenging configurations, which is relevant to scalability.

A notable finding from the literature is that, despite the high performance of commercial systems, few papers rigorously evaluate their use in truly large-scale or industrially challenging environments. Some research explores alternative low-cost multicamera systems combined with an extended Kalman filter (EKF) or portable near-infrared camera approaches \cite{Meyer2024DesignFilter, Lvov2023MobileOn-The-Go}, although these systems are typically validated only in small or medium-sized settings. Studies such as \cite{Hansen2022UAVSolution} demonstrate a comparison of fiducial-based SLAM benchmarking methods in large industrial-like spaces but do not fully address scalability, calibration complexity, or environmental robustness. In general, while commercial MoCap systems provide excellent precision, the literature reveals significant gaps in understanding how they perform and scale within expansive environments like aircraft hangars.

\subsection{Ultra-wideband Systems}
UWB localisation systems have become efficient methods for achieving accurate indoor positioning within complex, RF-challenging environments, such as industrial sites. Generally, these systems employ Time-of-Arrival (ToA) or Time-Difference-of-Arrival (TDoA) methods, which can deliver centimetre-level precision when conditions are optimal \cite{Zafari2019ATechnologies}. UWB systems operate by transmitting short-duration pulses over a wide frequency spectrum, enabling them to effectively mitigate multipath interference and signal fading, common challenges encountered by narrowband solutions in complex indoor environments.

Studies conducted in structured real-world environments, such as industrial plants \cite{Schroeer2018AScenarios}, confirm the strong performance of UWB in environments with metallic surfaces and multipath interference. UWB systems offer key advantages over vision-based alternatives, including robustness to occlusions and lighting variations, as well as reduced computational complexity in localisation pipelines due to lightweight signal processing requirements \cite{Fatima2022HighSafety}. However, the deployment of UWB systems also presents challenges, particularly for applications such as aircraft hangars. Multipath effects caused by reflective environments and dynamic obstructions (e.g., moving machinery or personnel) can degrade precision if not properly addressed. Several approaches have been proposed to mitigate these issues, such as machine learning-based corrections \cite{Karadeniz2020PreciseNodes} and multipath-assisted localisation techniques \cite{Wang2024Multipath-AssistedLearning}. Scalability is a relative strength; systems have been successfully deployed in facilities exceeding 1500 m\textsuperscript{2} \cite{Leugner2018LessonsApplication}, but the complexity of the setup increases with the size of the system.

A typical UWB localisation system consists of multiple fixed UWB anchors installed at known positions within the environment and UWB tags mounted on the object of interest (Figure \ref{fig:topologies}b). The system estimates the position of the tag by measuring the time of flight (ToF), time difference of arrival, or round-trip time (RTT) of UWB radio pulses exchanged between the tag and anchors. A localisation engine processes these timing measurements on an external computing unit to compute the tag's position through multilateration or filtering techniques such as Kalman or particle filters. Synchronisation between anchors, via wired or wireless methods, is crucial for accurate timing. UWB systems are often integrated with additional sensors such as inertial measurement unit (IMU), light detection and range (LiDAR) or vision systems to improve localisation robustness and accuracy, particularly in non-line-of-sight or multipath conditions. For large-scale, high-precision applications like aircraft hangars, UWB presents a strong foundational technology, provided that deployment challenges are managed through intelligent design and calibration strategies.

\begin{figure}
    \centering
    \includegraphics[width=0.5\linewidth]{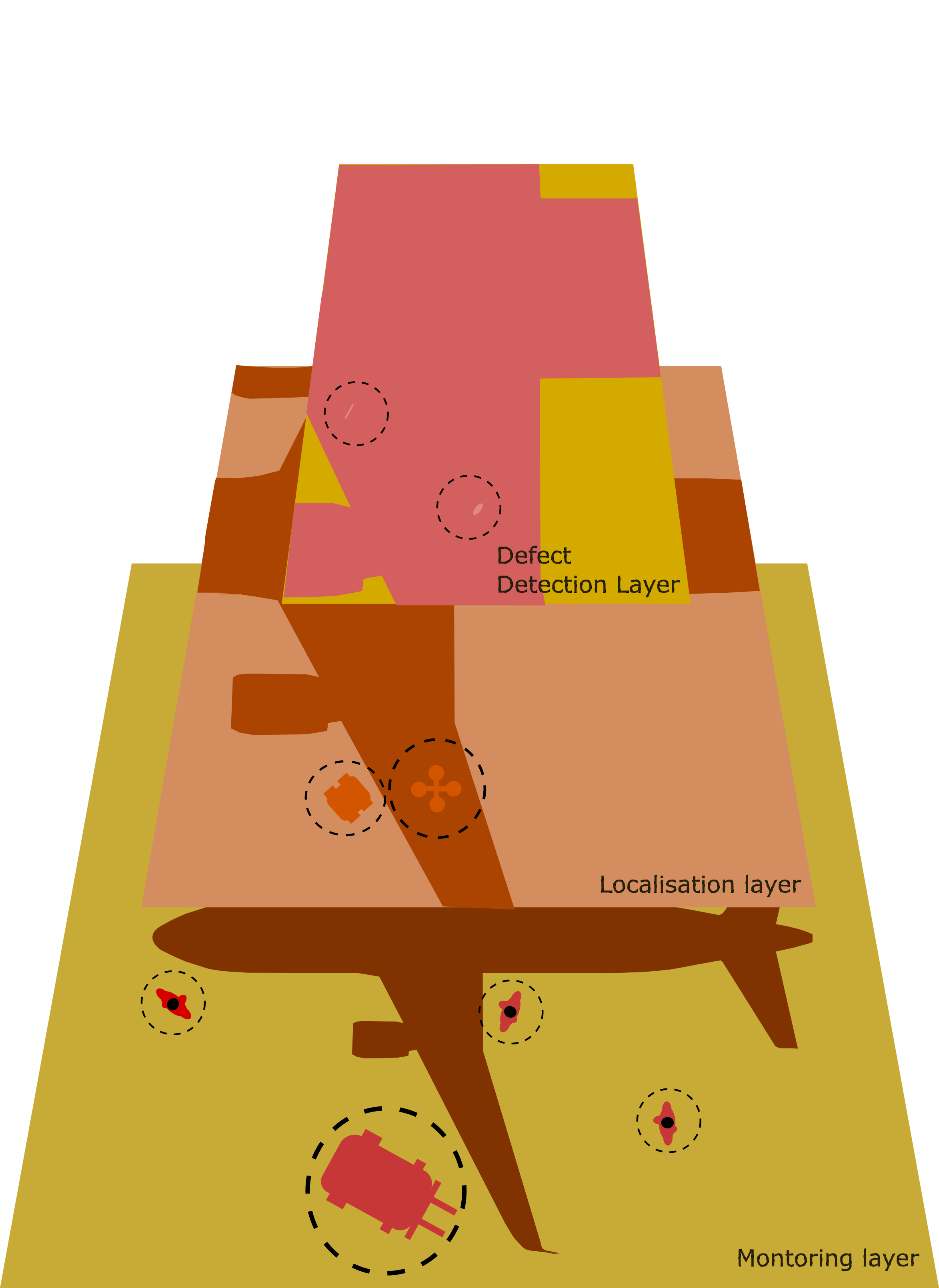}
    \caption{The camera system could be configured to operate in three modes: monitoring,  localisation, and defect detection.}
    \label{fig:pyramid}
\end{figure}

\subsection{Ceiling-Camera Vision}
Utilising a camera-based setup provides versatility in detecting and tracking objects of interest. The system parameters, including sensor dimensions and the lens's field of view, can be modified to suit the desired application. The system is simpler compared to the other approaches, as it consists of cameras, PoE network switches, and the computing unit that executes the required algorithms (Figure \ref{fig:topologies}c). As the primary functionality of the system relies on the algorithms, this allows the system to be repurposed depending on the target scenario. There are three modes that can be considered using the camera system. As illustrated in Figure \ref{fig:pyramid}, it can be used to locate robotic platforms, monitor assets, or detect defects. In the following subsections, for each mode, the typical lateral dimensions of the target objects, the height, and the velocity are defined. Every operational parameter is linked to the specifications of the camera. Table \ref{tab:objects-or-interest-specs} summarises the assumptions that were analysed in the following subsections.


\begin{figure}
    \centering
    \includegraphics[width=0.5\linewidth]{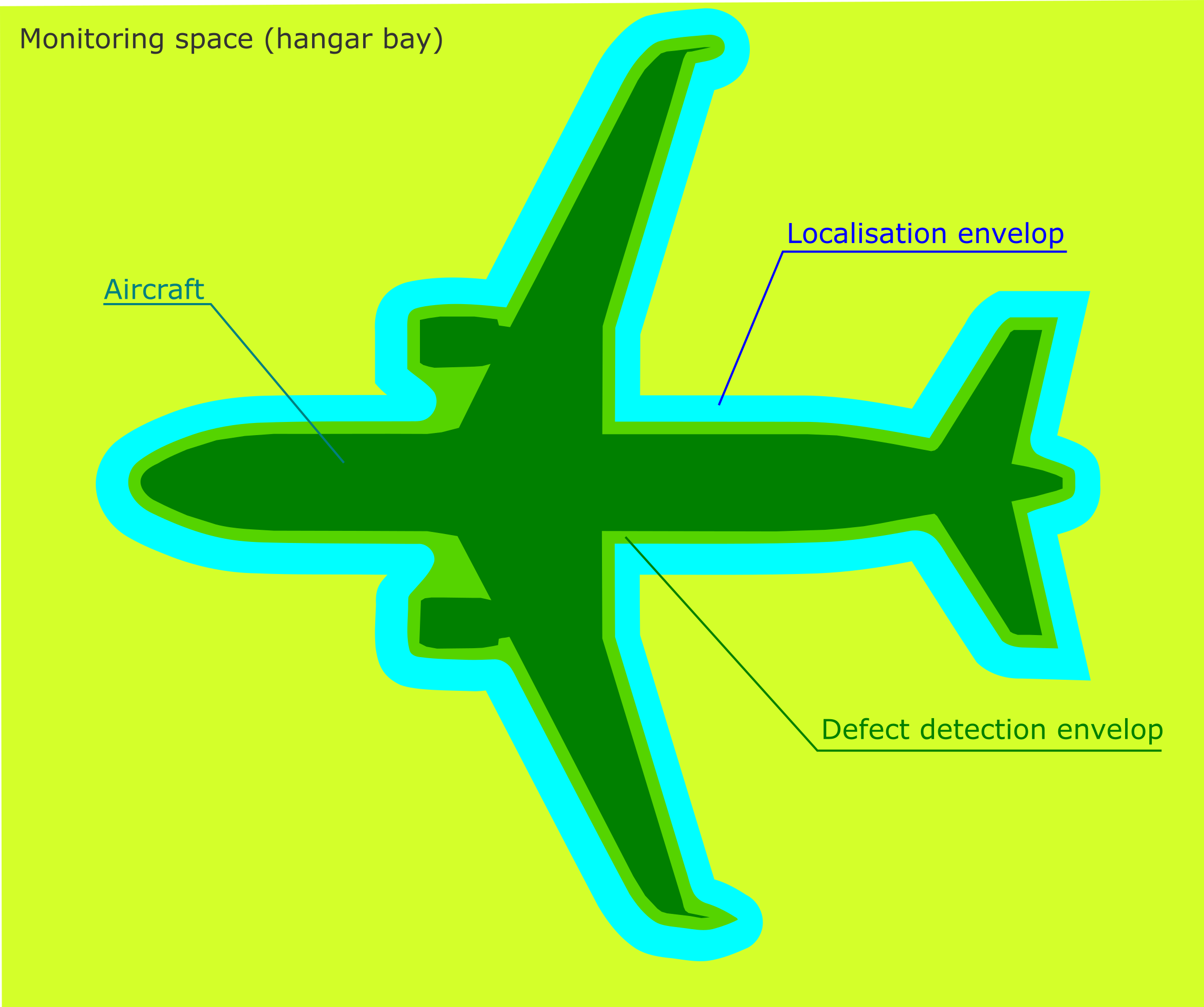}
    \caption{Different candidate modes of operation for the camera-based system.}
    \label{fig:space_considerations}
\end{figure}

\subsubsection{Drone and Ground Platform Localisation Mode}
Robotic platforms for inspection purposes can vary from aerial systems to ground units. The choice of platform is determined by the Nondestructive-testing (NDT) sensors and the inspection target. Drones are well-suited for inspecting vertical and horizontal stabilisers, as well as the upper sections of the fuselage and wings. Meanwhile, landing gears and areas beneath the wings and fuselage are more suitable for ground-based robotic platforms. Depending on the demands and budget, options such as wheeled, tracked, or even quadruped systems can be explored.

There are many approaches available for drones classification based on size, mass, and altitude. In this study, the focus is on the frame size, which is often referred to as the diameter. The range of mini quad-copters usually ranges between 250 and 1000 mm, with 500 mm being a common choice depending on the application \cite{Tatale2018Quadcopter:Testing}. An acceptable presumption for the drone's speed during inspection activities might fall between 0.5 and 1.5 m/s. Regarding wheeled robots, representative cases can be considered Husarion's Panther and Clearpath's Husky, with external dimensions (L x W x H) 810 x 850 x 370 mm and 990 x 698 x 372 mm, respectively \cite{HusarionPantherSpecifications, ClearpathRoboticsHuskySpecifications}. The most popular quadruped, which is the Boston Dynamics Spot robot, has external dimensions of 1100 x 500 x 610 mm \cite{BostonDynamicsSpotSpecifications}. Considering these typical cases, assuming a drone with minimum lateral dimensions of 500 by 500 mm constitutes a valid assumption. Similarly to drones, it is reasonable to assume an inspection velocity ranging from 0.5 to 1.5 \si{\metre\per\second}. In the drone's localisation, the assumption is that the drone will fly 0.5 to 1.0 m above the aircraft surface, which translates to an altitude of 4.5 to 6.5 m. Regarding ground mobile platforms, it is acceptable to assume heights in the range between 380 mm and 600 mm. In terms of operational space, the focus is on covering the aircraft and the surrounding region. Figure \ref{fig:space_considerations} shows the aircraft along with the coverage envelope (blue indigo), which expands the boundary of the aircraft by 1 m. The focus in terms of distance from the ground is 4.0 to 7.0 mm for drones and 0.5 to 1 m for ground platforms.

\subsubsection{Asset and Personnel Tracking Mode}
\begin{figure}
    \centering
    \includegraphics[width=1\linewidth]{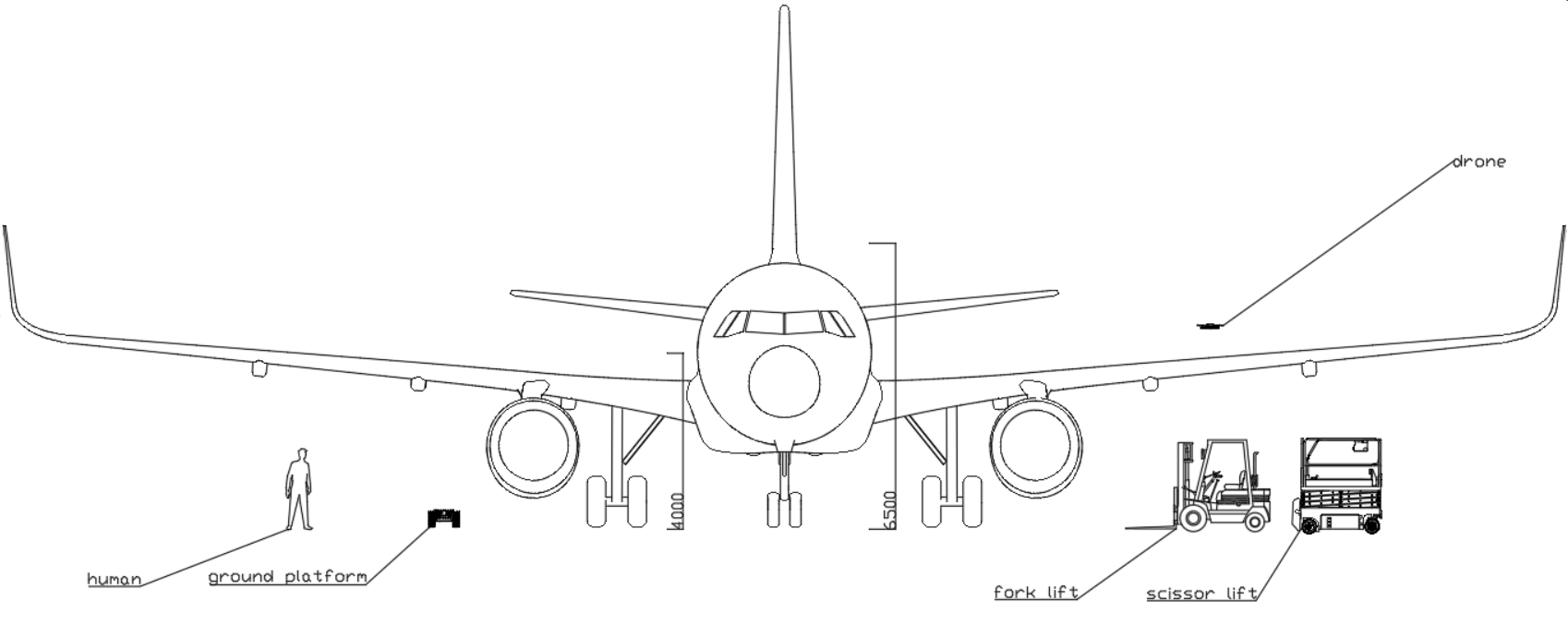}
    \caption{The front view of relevant assets inside a hangar in real scale (aircraft Airbus A320).}
    \label{fig:heights}
\end{figure}

Hangars are busy environments with a variety of ground support vehicles performing tasks during MRO operations. Some examples could be scissor lifts or cherry pickers (L x W) 2.4 x 1.1 m, forklifts 3.8 x 1.2 m, ground support units 2.8 x 1.7 m and tug tractors 4.5 x 2.0 m. This list is not exhaustive, but it contains the most frequent vehicles found inside the hangar. In different scale size, and not being a vehicle, tooling trolleys are also frequently around technicians performing maintenance with typical lateral dimensions 950 x 460 mm. A significant factor to consider in the monitoring mode is the inclusion of humans. The analysis presented in \cite{Panero1992HumanStandards} suggests that humans can be represented as ellipses, where the minor axis corresponds to body depth and the major axis corresponds to shoulder width. The shoulder width is estimated to be 610 mm, while the body depth is approximately 457 mm. Typical indoor speed limits for support vehicles moving inside the hangar range between 0.9 and 1.8 \si{\metre\per\second}, when humans are present and driving in free space, respectively \cite{ForkliftsafetyWhatOperations, TransmonEngineeringLtdManagingSpeeds}. In humans, walking speed is an essential gait parameter that indicates the functional state and overall health of a person. This metric is typically assessed by the distance covered over a period of time, with speeds varying from a leisurely 0.82 \si{\metre\per\second} to a brisk 1.72 \si{\metre\per\second} \cite{Murtagh2021OutdoorMeta-analysis}. In terms of height, the lower and upper surfaces of the support vehicles are within the range of 1200 mm and 2000 mm (Figure \ref{fig:heights}). If humans are included, according to \cite{Bentham2016AHeight}, the average height of an adult male could be approximated to 1700 mm, which falls within the aforementioned range. The operational space is the area surrounding the aircraft where maintenance tasks are performed (highlighted in yellow in Figure \ref{fig:space_considerations}). According to building manufacturers \cite{SpanTechNarrowHangars}, the standard size of a single-bay narrow-body maintenance space varies from 40 x 50 metres to 60 x 60 metres. The distance from the ground varies depending on the target; for vehicles, it is 2 metres, for humans, approximately 1.7 metres, and for tool trolleys, 1 metre.

\subsubsection{Surface-Defect Detection Mode}
Commercial aircraft surfaces are prone to a range of defects that can occur during manufacturing, use, and maintenance. These defects, which range from tiny corrosion pits to noticeable dents and cracks, significantly affect the structural integrity and safety of the aircraft. 

Among the most frequently encountered surface defects on aircraft are dents. These defects arise from various sources, including hail impacts, tooling mishaps, foreign object debris (FOD), or cracks due to bird collisions or debris ingestion. The dimensions of the dents depend on the energy of the impact, the properties of the material, and the environmental conditions. Hail damage produces dents with depths proportional to the diameter of the hailstone and the angle of impact \cite{Hayduk1973HailSurfaces}. Chen et al. \cite{Chen2014InspectionDamage} conducted an analysis of damage records spanning more than a decade and determined that the usual sizes range from 38 to 50 mm. In the same work, they mentioned that manual visual inspection usually identifies dents greater than 10 mm in lateral dimensions.

The deterioration of paint and coatings on commercial aircraft surfaces may manifest itself in multiple ways, such as peeling, chipping, cracking, damage from ultraviolet light and oxidation. These issues not only diminish the aircraft's visual appeal but may also impact its structural soundness and aerodynamic performance. Peeling can vary significantly in size, from a few square centimetres to more extensive degradation if not addressed promptly. The coating thickness for aircraft paint typically ranges from 0.1 mm to 0.5 mm \cite{Moupfouma2013AircraftDamages}. Physical damage can result from lightning strikes at their impact sites. As reported by \cite{Fisher1982Lightning80}, the typical evidence was the presence of molten metal that had resolidified, appearing as spots with diameters ranging from 1 to 10 mm. Factors such as UV exposure, temperature changes, and mechanical stress can cause the formation of coating cracks. According to \cite{Rojas2015InnovativeCracks}, these cracks can differ in length, with visual inspections aiming at detecting cracks of at least 12.7 mm. In terms of height, for narrow-body aircraft, the upper part of the wings and fuselage is within the range of 4.0 to 6.0 m.

Summarising the key aspects of the earlier analysis, it is safe to assume that an optical system capable of detecting defects between 10 and 50 mm may prove to be a valuable aid in general visual inspections (GVI). In defect detection mode, there is no need to consider velocity since the aircraft is static. This type of inspection is typically performed at regular intervals, but can also be arranged when a particular issue is suspected \cite{Baaran2009StudyStructures}.

In terms of space, the focus is strictly on the aircraft (in green). The light green area surrounding it represents a tolerance margin that accommodates any misalignment in the final placement of the aircraft in the bay (Figure \ref{fig:space_considerations}). This zone could be considered 0.5 m. The focus in terms of height is between 4 and 6.5 m, covering the wings and the upper part of the fuselage. 

\begin{table}[ht]
    \renewcommand{\arraystretch}{2} 
    \setlength{\tabcolsep}{4pt} 
    \caption{Summary of the modes of operation, the typical dimensions of the target objects and their velocity.}
    \label{tab:objects-or-interest-specs}
    \centering
    \resizebox{\linewidth}{!}{
    \begin{tabular}{@{}p{3cm} p{5cm} c c c@{}}
        \toprule
        \textbf{Mode} &\textbf{Objects of interest} &\textbf{Typical L x W (m)} &\textbf{Typical H (m)} &\textbf{Velocity (\si{\metre\per\second}})\\
        \midrule
        Localisation & Drones & 0.5 x 0.5 & 4.5 - 6.5 & 0.5 - 1.5\\ \hline
        Localisation & Wheeled platforms and quadrupeds & 0.8 x 0.5 & 0.38 - 0.6 & 0.5 - 1.5\\ \hline
        Monitoring & Scissor lifts, cherry pickers, forklifts, tugs  & 2.4 x 1.1 & 1.2 - 2.0 & 0.9 - 1.8\\ \hline
        Monitoring & Scissor lifts, cherry pickers, forklifts, tugs, humans and trolleys & 0.61 x 0.46 & 1.0 - 2.0 & 0.8 - 1.7\\ \hline
        Defect Detection & Small defects such as cracks and lightning strikes & 0.1 x 0.1 & not applicable & not applicable\\ \hline
        Defect Detection & Medium defects such as dents and peeled paint & 0.4 x 0.4 & not applicable & not applicable\\ \hline
        \bottomrule
    \end{tabular}
    }
\end{table}

\subsection {Comparative analysis of the enabling technologies}
This section brings together the assessments of MoCap, UWB, and vision systems relying on cameras discussed earlier into a streamlined comparison matrix. Table \ref{tab:systems-comparison} contrasts their fundamental abilities, infrastructure requirements, and specific trade-offs, providing a clear overview of the performance of each technology within the confines of a large aircraft hangar. 

{
\setlist[itemize]{nosep, leftmargin=*}
\begin{longtable}{@{}p{2.5cm} p{3.5cm} p{3.5cm} p{3.5cm}@{}}
    \caption{Key characteristics of the three localisation and monitoring systems} 
    \label{tab:systems-comparison} 
    \\
    
    \toprule
    \textbf{Evaluation aspect} & 
    \textbf{MoCap} & 
    \textbf{UWB} & 
    \textbf{Camera-based} \\
    \midrule
    \endfirsthead

    \toprule
    \textbf{Evaluation aspect} & 
    \textbf{MoCap} & 
    \textbf{UWB} & 
    \textbf{Camera-based} \\
    \midrule
    \endhead

    \midrule
    \multicolumn{4}{r}{\small\em Continued on next page} \\
    \midrule
    \endfoot

    \bottomrule
    \endlastfoot
        Typical accuracy and rate & 
        Sub-millimetre static ($\approx$0.15~mm) and $<$ 2~mm dynamic errors at high update rates &
        Centimetre-level position when conditions are good &
        Varies with algorithm; not yet benchmarked, still a research topic 
        \\

        Core strengths & 
        \begin{itemize}
            \item Highest precision available
            \item Provides full 6-DoF pose
            \item Mature commercial ecosystems
        \end{itemize} & 
        \begin{itemize}
            \item Robust to occlusion and lighting changes
            \item Low on-board processing load
            \item Proven in $>$1500 ~m$^2$ industrial halls
        \end{itemize} &
        \begin{itemize}
            \item Hardware is simple (PoE cameras, switch, computing)
            \item Highly versatile—same hardware can localise robots, monitor assets or inspect defects by changing software
        \end{itemize} 
        \\
        
        Main limitations & 
        \begin{itemize}
            \item Very expensive hardware
            \item Needs clear line-of-sight to $\geq$~2 cameras per marker
            \item Occlusion and wide-area calibration are challenging in hangars
        \end{itemize} & 
        \begin{itemize}
            \item Position error rises in strong multipath and metal-rich zones; careful calibration needed
            \item Extra anchors could raise complexity as the area grows      
        \end{itemize} & 
        \begin{itemize}
            \item Performance is highly sensitive to illumination and viewpoint
            \item Heavy computation for deep-learning pipelines
            \item Not yet adopted or standardised; real-hangar robustness still unproven
        \end{itemize} 
        \\
        
        Infrastructure footprint &
        IR cameras synchronisation, network wiring, reflective markers attached to the object of interest and calibration rigs & 
        Fixed ceiling/wall anchors, network wiring, clock-sync links and tags on every asset &
        Ceiling-mounted industrial cameras, network wiring. 
        \\ 

        Scalability inside a large hangar &
        Precision typically drops when cameras are far apart; occlusion grows with wings and tooling & 
        Good coverage range, but anchor count and calibration effort rise with bay size &
        Theoretical scalability is high, yet coverage planning and algorithm speed must keep pace with camera count 
        \\
        
        \bottomrule
\end{longtable}
}

\section{Optimised and cost-effective camera selection and coverage mapping}
The preceding section discussed three potential systems for the Smart Hangar, namely MoCap, UWB, and camera-based solutions. Although the first two options are available on the market, the camera-based system remains an area of open research. This section formalises the algorithmic framework for the camera-based architecture. It addresses how to select a commercially available camera-lens combination and strategically position a minimal number of units to capture the entire area of interest with the necessary ground sampling distance, frame rate, and budget, depending on the operational mode (robot localisation, asset monitoring, or defect detection). The ensuing framework integrates classical pinhole projection, cost-weighted distortion heuristics, and binary set-cover optimisation to translate high-level operational needs into a practical camera deployment layout.

\begin{figure}[h!]
    \centering
    \includegraphics[width=1.0\linewidth]{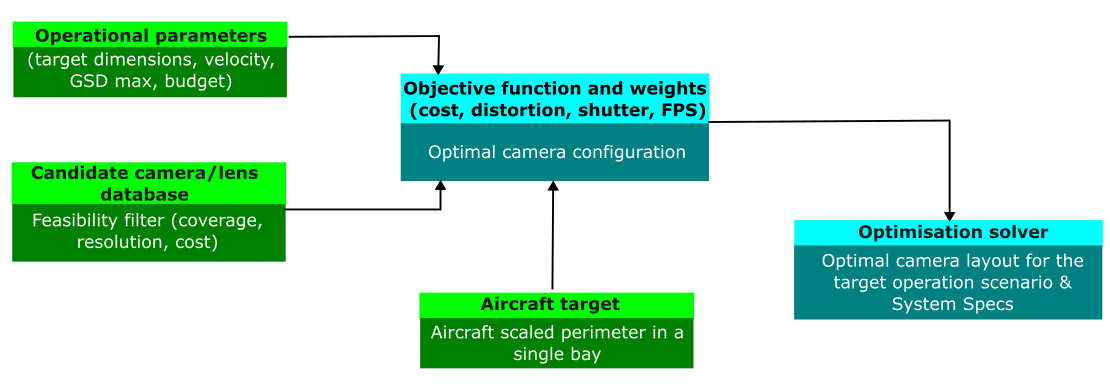}
    \caption{A high-level block diagram of the double-layer optimisation procedure for the optimal camera layout algorithm.}
    \label{fig:method_flow}
\end{figure}

The process is divided into four steps. Initially, the envelope of the target object, the working distance, and the maximum allowable ground sampling distance (GSD) are used to sift through a database of camera–lens pairs for initial feasibility. Second, an objective function balances hardware cost against optical distortion, shutter type, and frame-rate penalties, allowing each surviving pair to be ranked. In the third step, the field-of-view of the chosen pair is projected onto the hangar top-view, where a discretised grid of potential camera centres is created, and a Boolean visibility matrix is constructed. Finally, a mixed-integer linear approach addresses the resulting set-cover problem, determining both the number and positions of cameras needed to achieve full coverage with a specified overlap. Figure \ref{fig:method_flow} depicts a high-level block diagram of the solution.

At first, the algorithm identifies a feasible camera-lens pair to cover a target area at a specified working distance while respecting a maximum GSD, budget constraints, and an approximate lens distortion penalty. This approach follows the classic pinhole camera model in photogrammetry and computer vision. The premise involves using a singular camera lens setup to observe a flat surface located at a working distance of \textit{d} (in mm). The camera sensor has physical dimensions (S\textsubscript{w}, S\textsubscript{h}) (in mm), and the lens has a focal length \textit{f} (in mm). Under the pinhole camera approximation, the field of view (FoV) in each dimension can be calculated by projecting the sensor size onto the object plane at a distance \textit{d}:
\begin{equation}
    \text{FoV}_w = \frac{d \times S_w}{f}, \quad \text{FoV}_h = \frac{d \times S_h}{f}
    \label{eq:fov}
\end{equation}
The diagonal FoV is defined as:
\begin{equation}
    \text{FoV}_{\text{diag}} = \sqrt{(\text{FoV}_{w})^2 + (\text{FoV}_{h})^2}
    \label{eq:fov_diag}
\end{equation}
In photogrammetry, GSD represents the real-world size of one pixel in the captured image. If the camera resolution in the horizontal direction is R\textsubscript{w} pixels (width) and R\textsubscript{h} pixels (height), the GSD in each dimension is:
\begin{equation}
    \text{GSD}_w = \frac{\text{FoV}_w}{R_w}, \quad \text{GSD}_h = \frac{\text{FoV}_h}{R_h}
    \label{eq:gsd}
\end{equation}
To maintain adequate spatial resolution in the images, these values should remain below a user-defined maximum GSD limit, denoted as GSD\textsubscript{max}.
Although the camera specifications lack distortion coefficients, it remains possible to implement a distortion minimisation strategy using estimations and heuristic methods. An uncomplicated method is to presume that larger fields of view result in increased distortion. The corresponding equation is as follows:
\begin{equation}
    D = \frac{\text{FoV}_{\text{diag}}}{f}
\end{equation}
A larger diagonal FoV for a given focal length \textit{f} implies a higher distortion metric \textit{D}. Although simplistic, this metric provides a convenient scalar measure to compare different camera lens configurations.
The best combination of camera lenses for monitoring a target area is determined using a minimal optimisation algorithm. This algorithm considers factors such as working distance, desired coverage area, pixel density, and budget constraints. The primary objective is to reduce overall costs while meeting all operational needs. Assuming that the camera cost is C\textsubscript{cam} and the lens cost is C\textsubscript{lens}, the definition of the total cost is the following.
\begin{equation}
    C_{\text{total}} = C_{\text{cam}} + C_{\text{lens}}
    \label{eq:cost_equation}
\end{equation}
To select among feasible camera-lens pairs, a potential multi-objective function can be:
\begin{equation}
    O = C_{\text{total}} + \alpha D - \beta \cdot \text{GS} + \gamma \cdot \text{FPS penalty}
    \label{eq:placeholder}
\end{equation}
Using this approach, the aim is to rank feasible options by combining hardware cost, distortion, shutter type preference, and frame rate stability. The constants $\alpha$, $\beta$, and $\gamma$ allow tuning of trade-offs between distortion, shutter preference, and temporal resolution. More specifically, the parameter \textalpha{ } is a user-chosen weight to balance monetary cost versus the distortion penalty \textit{D}. In practice, the parameter \textalpha{ } may be chosen according to the desired level of penalisation for wide fields of view or lens distortion. The parameter $\beta$ assigns a cost-reduction bonus to configurations that include a global shutter sensor. This reflects the practical advantage of global shutters in avoiding motion blur during dynamic operations, making them more suitable for applications involving moving objects. The parameter $\gamma$ penalises configurations with non-ideal frame rates. Frame rates below 20 FPS may result in choppy visual streams, while rates above 50 FPS are often unnecessary and can increase bandwidth and processing load. This term helps to prioritise configurations that operate within a desirable temporal resolution range.

Before applying ranking configurations, a filtering step is applied to discard those that do not meet these fundamental criteria. This ensures that only operationally feasible combinations are passed on to the optimisation process, reducing computational overhead and improving the quality of the result.

In terms of constraints, each candidate configuration must satisfy the following:
\begin{itemize}
    \item Coverage: \begin{equation}
    \text{FoV}_w \geq W_{\text{target}}, \quad \text{FoV}_h \geq H_{\text{target}}
    \label{eq:field_of_view_constraints}
\end{equation} where W\textsubscript{target} and H\textsubscript{target} is the required coverage size (in mm).
    \item Resolution: \begin{equation}
    \text{GSD}_w \leq \text{GSD}_{\text{max}}, \quad \text{GSD}_h \leq \text{GSD}_{\text{max}}
    \label{eq:gsd_constraints}
\end{equation}
    \item Budget: \begin{equation}
    C_{\text{total}} \leq C_{\text{max}}
    \label{eq:capacity_constraint}
\end{equation} where C\textsubscript{max} is the maximum allowable budget (in British Pounds).
\end{itemize}

Once the desired dimensions for the camera sensor have been determined, the next step is to estimate the field of view (FoV) of the camera positioned on the ceiling to inspect the aircraft. A geometric model was utilized, taking into account the sensor dimensions and the lens's focal length. The angular field of view (in degrees) is calculated for both the width and height of the camera sensor using the following equation:

\begin{equation}
\text{FoV} = 2 \cdot \tan^{-1}\left(\frac{s}{2f}\right) \cdot \left(\frac{180}{\pi}\right)
\end{equation}

where
\begin{itemize}
    \item \textit{s} is the sensor dimension (width or height) in millimetres,
    \item \textit{f} is the focal length in millimetres
\end{itemize}

During the space allocation phase, accurately scaling the aircraft's shape to real dimensions is crucial. The two most popular Original Equipment Manufacturers (OEMs) provide online catalogues of their aircraft models, commonly called 3-View drawings, which are used for gate planning purposes \cite{Airbus2025AutoCADDrawings, TheBoeingCompany2025CADPurposes}. These drawings are available in Drawing eXchange Format (DXF). Unfortunately, the way these drawings are presented is inconsistent, not only between different OEMs but also among various models from the same manufacturer. The focus is primarily on the top view. However, the drawings include not only the outline of the aircraft but also other components such as windows, doors, landing gears, antennas, and more. As a result, the view requires manual preprocessing to remove unnecessary elements that add complexity and create unwanted artefacts.

In the initial approach, the script processed DXF files to extract, scale, and adjust geometric shapes corresponding to aircraft contours. It extracted boundary points from DXF entities such as lines, polylines, ellipses, and arcs, grouped them into contours, and arranged them into polygons. Utilising custom interpolation and grouping parameters, it managed geometric variations and generated a data structure comprising the perimeter points. However, the approach had several limitations. It depends heavily on the specific format of the DXF file and the accuracy of the input data, such as the grouping threshold for contours and the scaling factor. Additionally, the generated polygons may not always be closed or valid due to the reliance on point ordering and proximity-based grouping. The approach is challenging to generalise for different use cases or varying DXF formats and its handling of complex shapes.

An alternative approach that ensures the calculated polygon is closed and eliminates the need for DXF preprocessing involves creating a Scalable Vector Graphics (SVG) representation of the aircraft's top-view perimeter. This process consists of importing the 3-View drawing into a vector drawing application and manually sketching the perimeter polygon on a new layer over the image. Given that this is a singular action required only once for each aircraft type, it is not regarded as a burdensome or unsustainable activity within the proposed process flow.

As one of the potential cases, the exterior linework of a narrow-body aircraft was parsed from the SVG file as an ordered vertex list.
    \begin{equation}
        \mathcal{V} = \left\{ (x_i, y_i) \right\}_{i=1}^N
        \label{eq:liinework}
    \end{equation}
with \textit{N} the number of vertices.
Assuming that $L_{\mathrm{px}}$ is the pixel length of the fuselage,
    \begin{equation}
        \label{eq:Lpx}
        L_{\mathrm{px}}
             =\max_i x_i-\min_i x_i ,
    \end{equation}
and let the certified aircraft length be $L_{\mathrm{m}}=37.6\,\si{m}$.  
The metric scale factor is therefore
    \begin{equation}
        \label{eq:alpha}
        \alpha=\frac{L_{\mathrm{m}}}{L_{\mathrm{px}}}\; ,
        \qquad
        (x_i^{\ast},y_i^{\ast})=\bigl(\alpha x_i,\;\alpha y_i\bigr)\; .
    \end{equation}
The reference polygon can be defined by the scaled vertex as
    \[
        \label{eq:scaling}
        \mathcal P
           =\operatorname{poly}\!\bigl\{(x_i^{\ast},y_i^{\ast})\bigr\}\subset\mathbb R^{2}.
    \]
To guarantee safe stand‐off for cameras or for the aircraft model, an external inspection envelope was introduced,  
$\mathcal P$ is uniformly offset by a Minkowski sum with a radius  
$\delta=1.0\,\si{m}$:
    \begin{equation}
        \label{eq:buffer}
        \mathcal P^{+}
           =\bigl\{\,\mathbf p\in\mathbb R^{2}\mid
                    \operatorname{dist}(\mathbf p,\mathcal P)\le\delta\bigr\}.
    \end{equation}
This is like inflating the perimeter of the plane by a specific distance.
In the next stage, a discretisation of the region of interest occurs.
A regular square grid of spacing
    \begin{equation}
        \label{eq:delta}
        \Delta=\lambda\
    \end{equation}
is seeded over the bounding rectangle of $\mathcal P^{+}$. The parameter $\Delta$ defines the granularity, but also affects the computation time of the approach.  
Let
    \begin{equation}
        \label{eq:points}
        \mathbb G_{\Delta}=\{(x,y)\in\mathbb R^{2}\mid
              x=x_{\min}+k\Delta,\;
              y=y_{\min}+\ell\Delta\}
    \end{equation}
be that lattice.  
The set of target points is determined by a Boolean flag based on whether interest is directed towards the internal or external space of the aircraft, contingent on the specific scenario.

    \begin{equation}
        \label{eq:gridPoints}
        \mathbf G=
        \begin{cases}
              \{\,\mathbf g\in\mathbb G_{\Delta}\mid \mathbf g\in\mathcal P^{+}\}, & \text{internal coverage},\\[4pt]
              \{\,\mathbf g\in\mathbb G_{\Delta}\mid \mathbf g\notin\mathcal P^{+}\}, & \text{external coverage}.
        \end{cases}
    \end{equation}
    
The cameras are mounted at a hangar ceiling height of \textit{h} and possess horizontal and vertical fields of view. In a pinhole model, the ground footprint of one camera is the rectangle
$\bigl[\!-\tfrac{W}{2},\tfrac{W}{2}\bigr]\!\times\!\bigl[\!-\tfrac{L}{2},\tfrac{L}{2}\bigr]$ with

    \begin{equation}
        \label{eq:WL}
        W = 2h\tan\!\bigl(\theta_h/2\bigr),\qquad
        L = 2h\tan\!\bigl(\theta_v/2\bigr).
    \end{equation}
    
Depending on the target scenario, an overlap coefficient $\beta$ controls the spacing of the lattice of candidate camera centres,

    \begin{equation}
        \label{eq:latStep}
        \Delta_x = \beta W ,\qquad
        \Delta_y = \beta L .
    \end{equation}
    
The origin of this lattice is shifted so that it coincides with the centroid of $\mathcal P^{+}$, ensuring a approximately symmetric layout around the aircraft.  
Let \(
        \{\mathbf c_1,\dots,\mathbf c_m\}
    \)
denote all candidate locations generated within the enlarged boundary box. Then, for every point on the grid $\mathbf p_i\in\mathbf G$ and camera candidate $\mathbf c_j$, an indicator was defined as
    \[
    A_{ij}
       =\begin{cases}
          1, & \mathbf p_i\in \text{Cov}(\mathbf c_j),\\
          0, & \text{otherwise},
         \end{cases}
    \]
where
    \(
    \text{Cov}(\mathbf c_j)
       =\mathbf c_j+\bigl[\!-\tfrac{W}{2},\tfrac{W}{2}\bigr]\times\bigl[\!-\tfrac{L}{2},\tfrac{L}{2}\bigr].
    \)
At this stage, the Boolean matrix $\mathbf A\in\{0,1\}^{n\times m}$ compactly describes which cameras see which points. A binary decision variables
    \(
    x_j\in\{0,1\}
    \)
indicate whether camera $j$ is installed. Ultimately, the minimum-camera problem is the classical set cover formulated as a binary integer linear programme:
    \begin{align}
    \min_{\mathbf x\in\{0,1\}^m}\quad & 
          \sum_{j=1}^m x_j ,
          \label{eq:obj}\\[4pt]
    \text{subject to}\quad &
          \sum_{j=1}^m A_{ij}\,x_j \;\ge 1,
          \qquad i=1,\dots,n.
          \label{eq:covConstr}
    \end{align}
Equations \eqref{eq:obj}–\eqref{eq:covConstr} are implemented using the \texttt{PuLP} \cite{Mitchell2011PuLPPython} library with category \texttt{LpBinary} (Linear-programming of binary type) and solved using the CBC MILP (COIN-OR Branch-and-Cut – Mixed-Integer Linear Programming) solver engine \cite{ForrestCoin-or/Cbc}.  
The optimal index set
    \(
    \mathcal C^{\star}
          =\{\,j\mid x_j^{\star}=1\}
    \)
specifies both the number and positions of cameras required for full coverage of the target region defined in \eqref{eq:gridPoints}.
Switching the Boolean flag \textit{coverage\_inside} immediately toggles the optimisation target from internal inspection to external surveillance without altering the rest of the framework.

\section{Design-to-Cost Deployment Blueprints}
\begin{figure}
    \centering
    \includegraphics[width=1.0\linewidth]{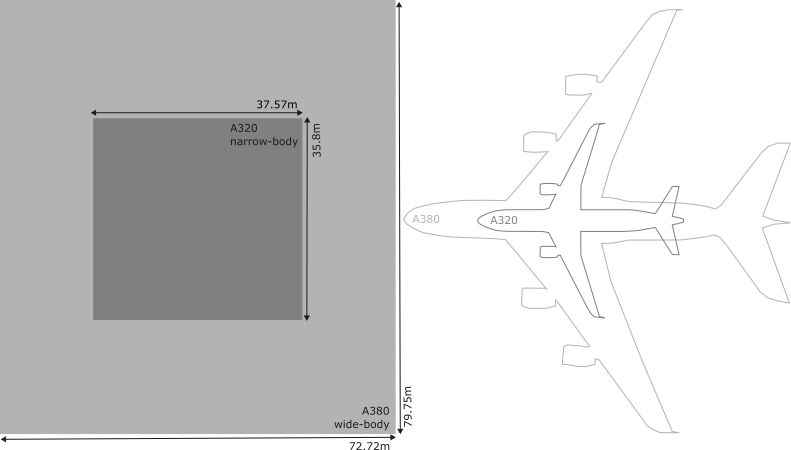}
    \caption{The size comparison between narrow-body (Airbus A320) and wide-body (A380) aircraft.}
    \label{fig:aircaft_comparison}
\end{figure}

Automated robotic inspection forms an integral part of the Smart Hangar concept, which utilises Industry 4.0 technologies to augment the abilities of human experts to perform necessary maintenance of the aircraft. The primary focus is on commercial aircraft that can be divided into two major categories. Narrow-body aircraft, also known as single-aisle aircraft, are the workhorses of short- to medium-haul routes. They typically have a single aisle, 2-3 seats on each side of the aisle, a capacity of 100 - 240 passengers, and a range of 3,000 - 4,000 nautical miles. Examples include Boeing 737 series and Airbus A320 family (A318, A319, A320, A321). The second category is the wide-body aircraft, or twin-aisle aircraft, which are designed for long-haul flights and have two aisles, 7-10 seats in economy class, capacity for 200 - 550 passengers, and a range of 5,000 - 8,000 nautical miles. Examples include Boeing 767, 777, 787 Dreamliner, and Airbus A330, A350, and A380. Figure \ref{fig:aircaft_comparison} illustrates the difference in size between the narrow-body and wide-body aircraft.

Taking into account the analysis presented in \cite{Statista2025AviationType} for the distribution of the fleet, the global operating commercial aircraft fleet in 2025 is dominated by narrow-body jets, which account for approximately 62\% of the total. Wide-body, regional, and turboprop jets are projected to comprise the remaining 38\% of the worldwide commercial aircraft fleet. 

The hangar dimensions are adjusted to the aircraft dimensions. For example, a hangar intended to host only narrow-body aircraft is required to have a door height of 12 m and an opening of 36 m. In the case of wide-body, the specifications are 24 in height and 75 m opening. In terms of ceiling height, the first case requires reaching 18 to 20 m, while the second case requires reaching 24 to 26 m. However, hangars are built with the intention of serving different types of aircraft, so they are usually able to host wide-body aircraft. In addition, the hangars can accommodate more than one aircraft simultaneously, featuring dedicated, predefined spaces called bays. The typical lateral dimensions \cite{SpanTechNarrowHangars} for a narrow-body aircraft hangar range from 40 x 50 m to 60 x 60 m.    

Recognising these standard features regarding hangars and fleet distribution in typical scenarios, it is generally assumed that a common scenario involves accommodating a narrow-body aircraft in a hangar with a roof height of about 22 to 24 metres and a bay area measuring around 40 to 50 metres. In the following sections, potential architectures are provided for each case. In the MoCap and UWB options, manufacturers provide suggestions for system deployments with an order-of-magnitude cost associated. For the camera-based system, a technical analysis is presented, based on the methodology outlined earlier, along with an estimated cost based on the components.

\subsection{MoCap Blueprint}
In the first approach, the concept of a hangar is approached purely as a 3D volume without considering the finer details. In the majority of MoCap systems, there are two modes for tracking objects: using active LEDs (active markers) or passive retro-reflective markers. One of the benefits of using the active is the camera range. With passive markers, the camera relies on the distance its own IR light can reach, ensuring sufficient power for the light to reflect back from a retro-reflective marker to the camera sensor. However, using active markers with the same camera, the range of that camera will increase. Since the hangar space is large, the first supplier recommended using active markers. A broad approximation for covering this area suggests using 160 cameras. The setup includes PoE switches, cables, a calibration wand, a synchronisation device, mounting accessories, software, a capture PC, and installation services. The estimated total cost is around £2,500,000.

\begin{figure}[h!]
    \centering 

    \begin{subfigure}[b]{0.5\textwidth} 
        \centering
        \includegraphics[width=\textwidth]{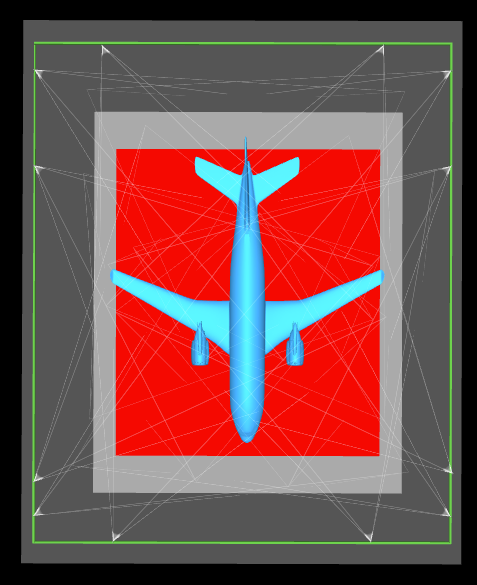}
    \end{subfigure}
    
    \vspace{1em} 

    \begin{subfigure}[b]{0.48\textwidth}
        \centering
        \includegraphics[width=\textwidth]{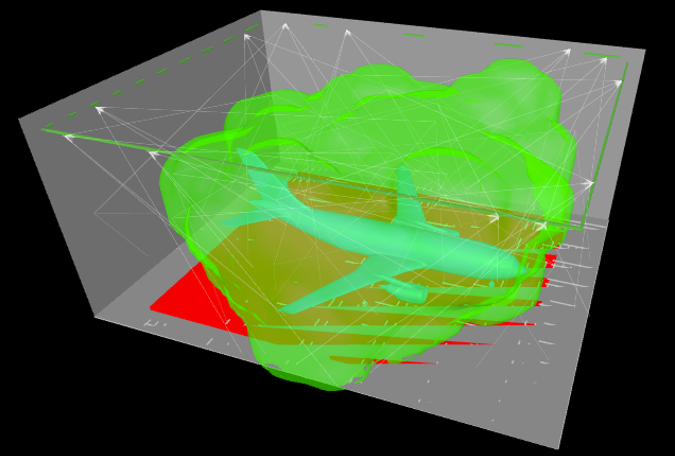}
    \end{subfigure}
    \hfill 
    \begin{subfigure}[b]{0.48\textwidth}
        \centering
        \includegraphics[width=\textwidth]{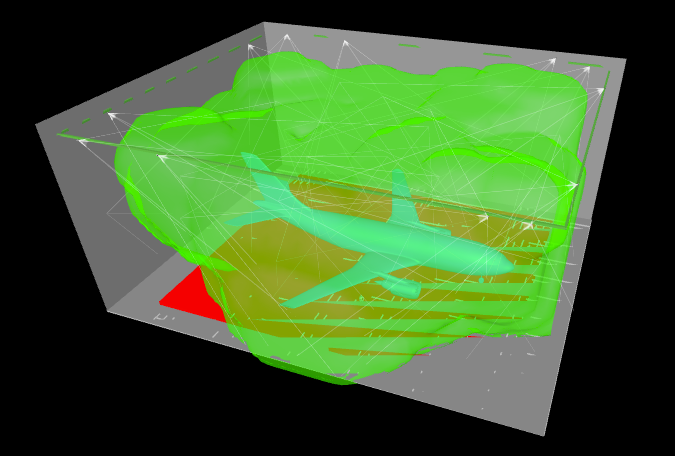}
    \end{subfigure}

    \caption{Top (a): MoCap system with 12 cameras set-up. Bottom left (b): MoCap system with 3-camera minimum tracking for each marker. Capture volume in shaded green. Bottom right (c): MoCap system with 2-camera minimum tracking for each marker.}
    \label{fig:qualisys}
\end{figure}

In the second approach, the system design is specifically tailored to optimise the performance and cost for the single-bay hangar scenario. The camera configuration involves installing 12 cameras at a height of 22 m, pointing downward (Figure \ref{fig:qualisys}a). Side or low-position cameras are excluded, as they are impractical for hangars with multiple bays. It is possible that it may need a few more cameras (perhaps increasing to a 16-camera system) because of occlusion around the wings and fuselage, but from lab simulations the proposed set-up offers a good estimate. The 12 cameras effectively cover a distance of about 32 m for a 19-mm marker. The suggested camera model is IP67, with a maximum capturing distance of 40 m and a 3D resolution of 0.04 mm. Figure \ref{fig:qualisys}b shows the capture volume, which demands that each marker be captured by three cameras. In reality, the system can track markers over a broader area than this, although a lower marker resolution might result in occasional tracking loss. In a MoCap system, tracking each marker requires the use of at least two cameras. Figure \ref{fig:qualisys}c illustrates the capture volume with two-camera tracking, where a larger area is covered toward the nose and tail. In practice, increasing the range of the camera does not provide much benefit. In addition, the size of the marker and the camera range increase proportionally, and markers of 30, 40, and 50 mm are commercially available, depending on the size of the robotic platform selected for the inspection. The estimated cost of this system is within the range of £180,000 to £200,000.

\subsection{UWB Blueprint}
Within the UWB coverage of the hangar, the extent of the necessary infrastructure is determined by the specific tracking applications used and the efficiency required in regions with significant obstructions, such as under the wings. In scenarios involving localisation and monitoring, such as the tracking of drones and ground platforms, the process is considerably simpler compared to tracking certain assets (e.g., tools). This is because it typically occurs above or between highly obstructed areas, avoiding locations like under wings. Considering the initial assumption of a single bay maintenance zone sized between 40 x 50 m and 60 x 60 m (resulting in a volume from 2000 to 3600 m\textsuperscript{2}) and noting that the application occurs in fairly open settings, a prudent estimate is to account for 10 to 25 anchors to achieve adequate coverage. For comparison, when applications involve tracking under the wing and fuselage, and around the engines, more sensors, possibly ranging from 20 to 40, might be used.

\begin{figure}
    \centering
    \includegraphics[width=0.75\linewidth]{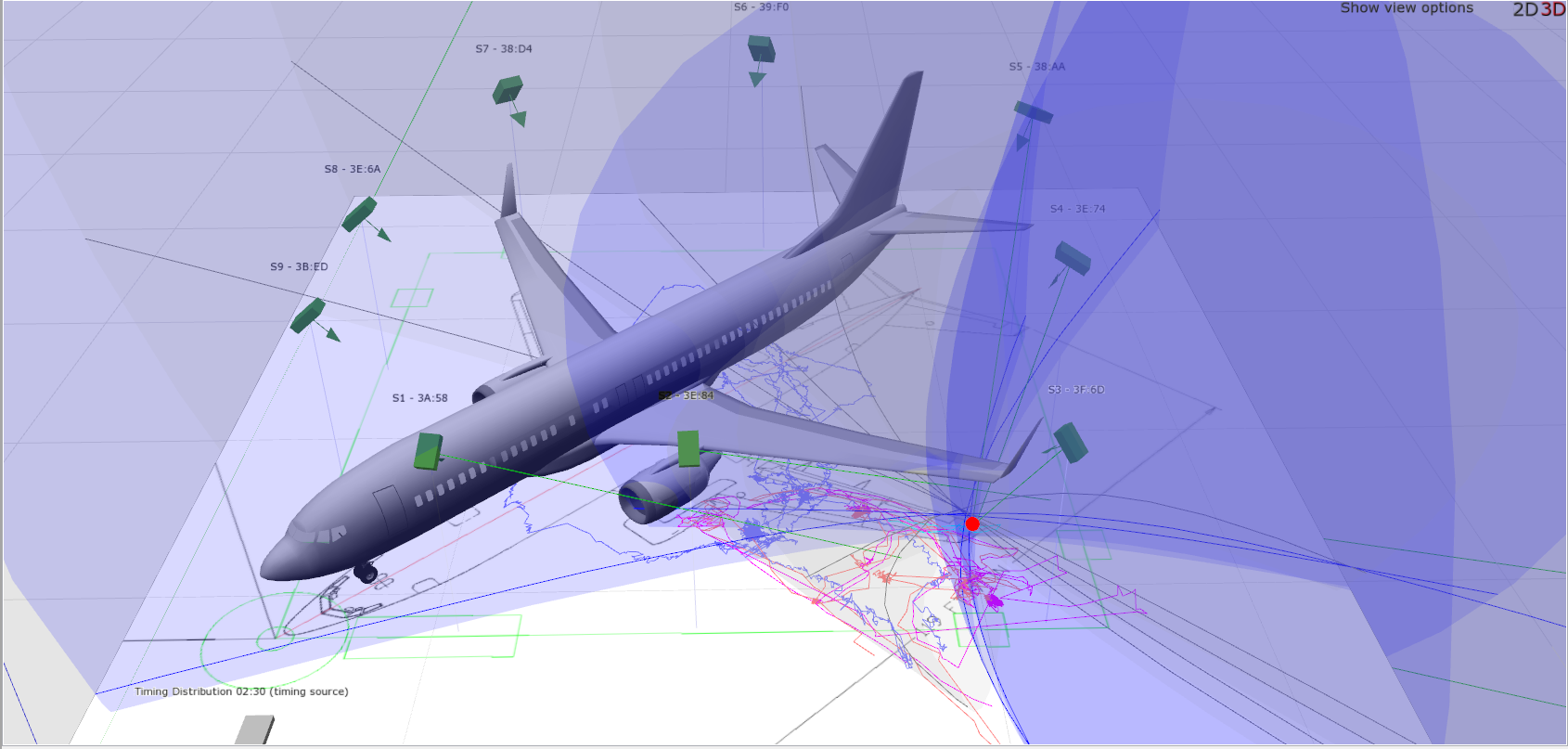}
    \caption{The UWB system installed at Cranfield's DARTeC Smart Hangar.}
    \label{fig:uwb-system}
\end{figure}

On the implementation side, power and network connectivity are required for the installation of each anchor. The most common practice involves using a Cat5e UTP Ethernet cable to connect each sensor location from a control room, which needs to be within 100 m of the sensor mounting location to comply with Ethernet networking limits. Given the size of the hangar, this likely necessitates equipment placement in two control rooms, each equipped with a Timing Distribution Unit for precise time synchronisation of the anchors and a PoE networking switch to provide power and networking to each sensor. The cost associated with running each cable can vary significantly depending on the location. It is largely determined by whether the cabling is installed during the construction of the hangar or after, whether the installation after construction occurs while the space is active (it is considerably more expensive if the space is active) and the geographical location where this activity occurs. Lastly, the cost of actually mounting the sensor on its bracket at the end of the cable needs to be included.

In order for the system to be completely functional, every object of importance must have a tag installed. A wide range of options is available regarding size, power consumption, and update rate. As an illustration, a small tag may feature dimensions of 46 x 42 x 18 mm and weigh about 20 g, with a battery life that exceeds 10 years while transmitting continuously at an update rate of 1 Hz. For drone tracking applications, the tags can be configured to transmit at a higher rate (e.g. 10 Hz), wherein the battery life might be approximately 12-15 months. These tags are priced at about £40 each. Another option is to incorporate a tag module into the drone or ground platform. This module draws its power from an external source, such as the drone's power system, eliminating the need for an internal battery that could run out. Naturally, the drone itself is periodically recharged. Without a battery restriction, the tag module can achieve considerably higher update rates, such as 50 Hz.

The UWB infrastructure is capable of monitoring thousands of tags within its coverage area and can deliver aggregate update rates of several thousand updates each second. The resulting data stream, composed of (ID, x, y, z, timestamp), is available for transmission to other applications. Although system suppliers offer various applications, there is also the possibility of transmitting the data as UDP network data, allowing it to be used by third-party connected systems. A standard UWB system setup, comprising anchors, PoE switches, TDUs, cabling, and calibration, is approximately valued at £49,000. This estimate is a general figure that may vary depending on the number of anchors used. An example setup of a UWB system is installed at the Digital Aviation Research and Technology Centre (DARTeC) at Cranfield University, where nine anchors and two types of tags with 0.5 and 30 Hz frequencies have been installed (Figure \ref{fig:uwb-system}). 

\subsection{Camera Blueprint}
The camera-based system is not a commercially available product, so the candidate implementation is not based on the suggestion of any manufacturer. Depending on the application scenario, the algorithms presented in Section 3 suggest a candidate system that can perform either the robot localisation, monitoring, or defect detection task. In the beginning, extensive product research was performed in order to identify the commercially available products that can be used as data for the analytical approaches. The focus was on machine vision cameras and not CCTV systems, as the former allows full control over defining the required characteristics without any compromises. 

In system design, camera specifications are crucial for optimal system performance. Beyond selecting the right camera and lens combination, it is essential to consider cost-effectiveness and the need to depict the object of interest at a suitable size for the deep learning component. The chosen camera features that were used as criteria for the analysis include sensor size and format, resolution, shutter type, pixel dimensions, frame rate, connectivity speed, and cost (Table \ref{tab:camera-specs}). For the lens, the most important characteristic was the focal length (Table \ref{tab:lenses-specs}). The compatibility between the camera and the lens was ensured as both supports the C-mount interface. Moreover, cameras come with two types of connection interface: USB-3 and PoE. For the analysis, PoE was chosen for all cameras, mainly due to its convenient installation on the hangar ceiling, and secondarily for its connection speed.

Each of the camera characteristics affects the specific design aspect that is captured in the proposed methodology. The specifications of the camera sensor and focal length, combined with the distance from the object of interest and its typical size, define the appropriate selection of the camera and lens of the system. In addition, the typical moving velocity of the object of interest is related to the frame rate to produce an adequate rate of image sequence to follow the motion. An additional characteristic that affects the performance of the system is the GigE connectivity interface, especially in terms of data transfer rate, latency, and, to a greater extent, the overall system throughput. The maximum theoretical data transfer rate of a standard GigE interface is 1 gigabit per second, which is equal to 125 megabytes per second. This determines the maximum frame rate and resolution that a camera can output. So, for example, a 2MP camera outputs uncompressed 8-bit images at 60 fps x 120 MB/s, which is close to the GigE limit. In general, GigE has higher latency compared to USB 3.0 interfaces. This potentially can affect real-time applications where an immediate response is needed (e.g., tracking). However, latency is often acceptable for many vision-based monitoring systems, especially if buffered. The last characteristic that was taken into consideration was the type of shutter. The global shutter camera takes a complete image in a single instance, as every pixel is exposed at the same time. There is no motion distortion, so fast-moving objects are captured without skew or wobble. This is the ideal case for precise measurements, which is useful in robotics, industrial vision, and 3D reconstruction. They can support easier synchronisation with external sensors like LiDAR or IMU, because the timestamp is consistent across the whole image. However, the rolling-shutter camera captures the image line-by-line from top to bottom (or side to side), over a short time interval. Because of this mode of operation, the image contains motion artefacts, which can cause skew, wobble, or partial exposure when objects move quickly or the camera moves during capture. In addition, it is harder to sync since different parts of the image correspond to slightly different times. In the algorithm analysis, all trade-offs are documented and are balanced by the input from the user, which is incorporated into the cost equation of the optimisation phase as weight parameters.

The camera feeds are forwarded to the computing unit via an Ethernet switch. To facilitate a simpler installation and reduce potential failure points, a PoE switch is suggested instead of using the power supply for each camera separately. When using high-resolution GigE cameras, it is recommended that Ethernet switches and Network Interface Cards (NICs) be used to support jumbo frames. In general, in Ethernet networking, a standard frame has a Maximum Transmission Unit (MTU) of 1500 bytes. Jumbo frames extend this limit, typically allowing up to 9000 bytes per frame. This means that more data can be transmitted in a single packet, reducing the number of packets needed for large data transfers. The advantages of utilising Jumbo frames include:
\begin{itemize}
    \item Reduced CPU load: Fewer packets mean fewer interrupts for the CPU to handle, lowering processing overhead.
    \item Improved Throughput: Larger frames can carry more data, enhancing overall network efficiency.
    \item Decreased Packet Loss: With fewer packets on the network, the chance of collisions and packet loss diminishes.
\end{itemize}
There are many commercial options available that satisfy these requirements. For example, the Mikrotik cloud router switch (CRS328-24P-4S+RM), which offers 24 PoE gigabit ports, with 4 SPF (Small Form-factor Pluggable) ports to connect fibre optics of different types and speeds. This specific device supports 10218-byte jumbo frames. A representative cost of this device is £417 (price captured from Amazon.co.uk on the 27th of April 2025).

\begin{longtable}{@{} p{1.5cm} p{1.5cm} p{1.5cm} p{1cm} p{1cm} p{1cm} p{1cm} p{0.8cm} p{0.8cm} p{1cm}@{}}
    \caption{The table presents a large selection of PoE cameras. The information was gathered from the Edmund Optics website, and the prices were captured on the 30th of March 2025.} 
    \label{tab:camera-specs} \\
    \toprule
    \textbf{Brand} & 
    \textbf{Sensor (mm)} & 
    \textbf{Resolution (px)} & 
    \textbf{Format (inches)} & 
    \textbf{Mpixels} & 
    \textbf{Shutter} & 
    \textbf{Pixel size (µm)} & 
    \textbf{Frame rate} & 
    \textbf{GigE} & 
    \textbf{Cost (£)} \\
    \midrule
    \endfirsthead

    \toprule
    \textbf{Brand} & 
    \textbf{Sensor (mm)} & 
    \textbf{Resolution (px)} & 
    \textbf{Format (inches)} & 
    \textbf{Mpixels} & 
    \textbf{Shutter} & 
    \textbf{Pixel size (µm)} & 
    \textbf{Frame rate} & 
    \textbf{GigE} & 
    \textbf{Cost (£)} \\
    \midrule
    \endhead

    \midrule
    \multicolumn{10}{r}{\small\em Continued on next page} \\
    \midrule
    \endfoot

    \bottomrule
    \endlastfoot
        Basler(1)&11.25x7.03&1920x1200&1/1.2''&2.3&Global&5.86&42.0&1.0&535 \\
        Basler(2)&9.22x5.76&1920x1200&2/3''&2.3&Global&4.8&50.0&1.0&612 \\ 
        Basler(3)&11.25x7.03&1920x1200&1/1.2''&2.3&Global&5.68&50.0&1.0&1032 \\ 
        Basler(4)&7.07x5.3&2048x1536&1/1.8''&3&Global&3.45&36.0&1.0&522 \\ 
        Basler(5)&11.26x11.26&2048x2028&1''&4.2&Global&5.5&25.0&1.0&1989 \\ 
        Basler(6)&5.7x4.28&2592x1944&1/2.5'' &5&Rolling&2.2&14.0&1.0&497 \\ 
        Basler(7)&12.43x9.83&2590x2048&1''&5&Global&4.8&5.0&1.0&943 \\ 
        Basler(8)&6.44x4.62&3856x2764&1/2.3'' &10.6&Rolling&1.67&10.0&1.0&544 \\ 
        Basler(9)&7.41x4.95&3088x2064&1/1.8''&6&Rolling&2.4&16.0&1.0&385 \\ 
        Basler(10)&7.44x5.62&4024x3036&1/1.7''&12&Rolling&1.85&8.0&1.0&463 \\ 
        Basler(11)&14.13x7.45&4096x2160&1''&9&Global&3.45&12.0&1.0&1487 \\ 
        Basler(12)&14.13x10.35&4096x3000&1.1''&12&Global&3.45&8.0&1.0&1819 \\ 
        Basler(13)&13.13x8.76&5472x3648&1''&20&Rolling&2.4&5.0&1.0&654 \\ 
        Basler(14)&6.6x4.1&1920x1200&1/2.3''&2.3&Global&3.45&51.0&1.0&332 \\ 
        Basler(14)&5.2x3.9&2592x1944&1/2.8''&5&Rolling&2.0&22.0&1.0&279 \\ 
        Basler(15)&7.7x4.3&3840x2160&1/1.8''&8.3&Rolling&2.0&13.0&1.0&414 \\ 
        Basler(16)&11.2x8.2&4096x3000&1/1.1’’&12.3&Global&2.74&44.0&1.0&1827 \\ 
        Basler(17)&14.6x8.3&5320x3032&1.1''&16.1&Global&2.74&34.0&1.0&1989 \\ 
        Basler(18)&12.3x12.3&4504x4504&1.1''&20.2&Global&2.74&27.0&1.0&2329 \\ 
        Basler(19)&14.6x12.6&5328x4608&1.2''&24.4&Global&2.74&22.0&1.0&2851 \\ 
        Lucid(1)&7.37x4.92&3072x2048&1/1.8''&6.3&Rolling&2.4&42.8&2.5&436 \\ 
        Lucid(2)&14.44x9.9&3208x2200&1.1’’&7.1&Global&4.5&35.0&2.5&1270 \\ 
        Lucid(3)&8.57x7.17&2448x2048&2/3’’&5.0&Global&3.45&35.6&2.5&697 \\ 
        Lucid(4)&7.78x7.78&2840x2840&2/3’’&8.1&Global&2.74&31.8&2.5&901 \\ 
        Lucid(5)&14.13x7.45&4096x2160&1’’&8.9&Global&3.45&13.7&1.0&1041 \\ 
        Lucid(6)&14.58x8.31&5320x3032&1.1’’&16.2&Global&2.74&17.5&2.5&1445 \\ 
        Lucid(7)&14.13x10.35&4096x3000&1.1’’&12.3&Global&3.45&22.6&2.5&1627 \\ 
        Lucid(8)&11.22x8.22&4096x3000&1/1.1’’&13.2&Global&2.74&21.7&2.5&1096 \\ 
        Lucid(9)&13.13x8.76&5472x3648&1’’&20&Rolling&2.4&6.0&1.0&522 \\ 
        Lucid(10)&13.13x8.76&5472x3648&1’’&20&Rolling&2.4&9.0&2.5&663 \\ 
        Lucid(11)&12.34x12.34&4504x4504&1.1’’&20.4&Global&2.74&13.5&2.5&1840 \\ 
        Lucid(12)&14.58x12.6&5320x4600&1.2’’&24.5&Global&2.74&11.3&2.5&2057 \\ 
        Flir(1)&14.13x7.45&4096x2160&1’’&8.9&Global&3.45&13.8&1.0&1389 \\ 
        Flir(2)&14.13x10.35&4096x3000&1.1''&12.3&Global&3.45&9.9&1.0&1717 \\ 
        Flir(3)&13.13x8.76&5472x3648&1’'&20.0&Rolling&2.4&6.1&1.0&595 \\ 
        Flir(4)&14.6x12.6&5320x460&4/3''&24.5&Global&2.74&5.0&1.0&1891 \\ 
        Allied(1)&8.5x7.1&2464x2056&2/3’’&5.1&Global&3.45&95.0&5.0&1377 \\ 
        Allied(2)&7.8x7.8&2848x2848&2/3’’&8.1&Global&2.74&59.0&5.0&1224 \\ 
        Allied(3)&7.4x5.6&4024x3036&1/1.1''&12.4&Global&2.74&39.0&5.0&1627 \\ 
        Allied(4)&7.44x6.61&4024x3036&1/1.7''&12.2&Global&1.85&41.0&5.0&760 \\ 
        Allied(5)&14.59x8.33&5328x3040&1.1''&16.2&Global&2.74&30.0&5.0&1899 \\ 
        Allied(6)&12.36x12.36&4512x4512&1.1''&20.4&Global&2.74&24.0&5.0&2231 \\
        Allied(7)&13.19x8.81&5496x3672&1’'&20.2&Rolling&2.4&25.0&5.0&1024 \\ 
        Allied(8)&14.6x12.63&5328x4608&1.2''&24.6&Global&2.74&20.0&5.0&2677 \\ 
        \bottomrule
\end{longtable}

\begin{longtable}{@{}p{2cm} p{7cm} p{2cm} p{1cm}@{}}
    \caption{The table presents a selection of lenses (C-Mount) compatible with the cameras presented in table \ref{tab:camera-specs}. The information was gathered from the Edmund Optics website, and the prices were captured on the 30th of March 2025.} \label{tab:lenses-specs} \\
    \toprule
    \textbf{Brand} & 
    \textbf{Type} & 
    \textbf{Focal length (mm)} & 
    \textbf{Cost (£)} \\
    \midrule
    \endfirsthead

    \toprule
    \textbf{Brand} & 
    \textbf{Type} & 
    \textbf{Focal length (mm)} & 
    \textbf{Cost (£)} \\
    \midrule
    \endhead

    \midrule
    \multicolumn{4}{r}{\small\em Continued on next page} \\
    \midrule
    \endfoot

    \bottomrule
    \endlastfoot
        Techspec(0)&4mm UC Series Fixed Focal Length Lens&4&208 \\
        Techspec(1)&6mm UC Series Fixed Focal Length Lens&6&221 \\ 
        Techspec(2)&8mm UC Series Fixed Focal Length Lens&8&221 \\ 
        Techspec(3)&12mm UC Series Fixed Focal Length Lens&12&221 \\ 
        Techspec(4)&18mm UC Series Fixed Focal Length Lens&18&518 \\ 
        Techspec(5)&25mm UC Series Fixed Focal Length Lens&25&518 \\ 
        Techspec(6)&35mm UC Series Fixed Focal Length Lens&35&518 \\ 
        Techspec(7)&50mm UC Series Fixed Focal Length Lens&50&518 \\ 
        Techspec(8)&75mm UC Series Fixed Focal Length Lens&75&561 \\ 
        Techspec(9)&100mm UC Series Fixed Focal Length Lens&100&561 \\ 
        \bottomrule
\end{longtable}
The candidate combinations of camera and lens are fed into the camera selection algorithm. Depending on the target scenario, the algorithm proposes five potential configurations that are illustrated in Table \ref{tab:camera selection}.

\begin{longtable}{@{}p{3.5cm} p{1.8cm} p{1.8cm} p{1.8cm} p{1.8cm} p{1.8cm}@{}}
    \caption{The table presents the camera-lens optimal combination per target scenario.} \label{tab:camera selection} \\
    \toprule
    \textbf{Specifications} & 
    \textbf{Defects\textsuperscript{1}} &
    \textbf{Localisation\textsuperscript{2}} & 
    \textbf{Localisation\textsuperscript{3}} & 
    \textbf{Monitoring\textsuperscript{4}} &
    \textbf{Monitoring\textsuperscript{5}} \\
    \midrule
    \endfirsthead

    \toprule
    \textbf{Specifications} & 
    \textbf{Defects\textsuperscript{1}} &
    \textbf{Localisation\textsuperscript{2}} & 
    \textbf{Localisation\textsuperscript{3}} & 
    \textbf{Monitoring\textsuperscript{4}} &
    \textbf{Monitoring\textsuperscript{5}} \\
    \midrule
    \endhead

    \midrule
    \multicolumn{6}{r}{\small\em Continued on next page} \\
    \midrule
    \endfoot

    \bottomrule
    \endlastfoot
        Target dimensions (mm)&40 x 40&500 x 500&800 x 500&2400 x 1100&600 x 450 \\
        Target dimensions (px)&45 x 45&83 x 83&114 x 71&490 x 224&121 x 91 \\
        Target area (m)&3 x 3&7 x 7&14 x 14&20 x 20&15 x 15 \\ 
        GSD (mm/pixel)&0.89&6.02&6.94&4.91&4.94 \\ 
        Camera &Allied(7)&Basler(1)&Lucid(2)&Lucid(11)&Allied(3) \\ 
        Lens &Techspec(7)&Techspec(4)&Techspec(3)&Techspec(3)&Techspec(2) \\ 
        
        \multicolumn{6}{l}{\textbf{Notes:}} \\
        \multicolumn{6}{l}{\textsuperscript{1} Defect detection refers to the ability to identify defects on the surface of the aircarft.} \\
        \multicolumn{6}{l}{\textsuperscript{2} Localisation refers to the process of determining the position of the drone.} \\
        \multicolumn{6}{l}{\textsuperscript{3} Localisation refers to the position detection for robotic platforms on the ground.} \\
        \multicolumn{6}{l}{\textsuperscript{4} Monitoring refers to tracking or monitoring moving vehicles.} \\
        \multicolumn{6}{l}{\textsuperscript{5} Monitoring involves tracking human movement or positions.} \\
\end{longtable}

\begin{figure}[h]
\centering
\begin{subfigure}{0.32\textwidth}
    \includegraphics[width=\linewidth]{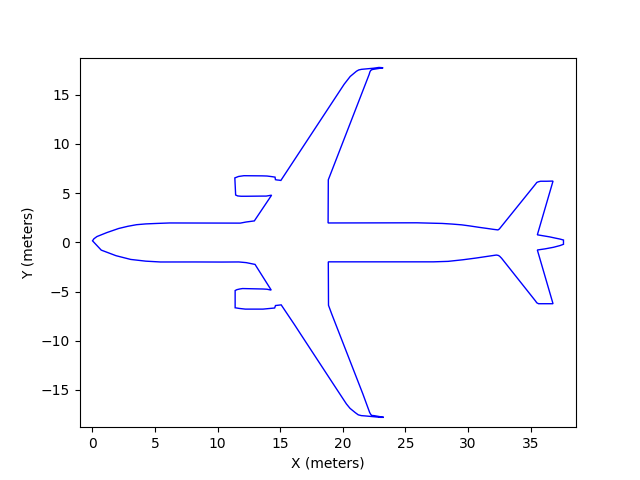}
    \caption{The Airbus A320 perimeter extraction.}
    \label{fig:a320-in-metres}
\end{subfigure}
\hfill
\begin{subfigure}{0.32\textwidth}
    \includegraphics[width=\linewidth]{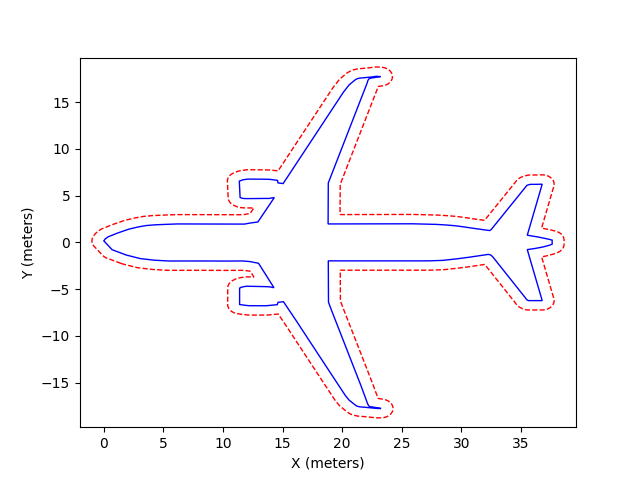}
    \caption{The Airbus A320 including the safe operation envelope (red dashed line).}
    \label{fig:a320-inflated}
\end{subfigure}
\hfill
\begin{subfigure}{0.32\textwidth}
    \includegraphics[width=\linewidth]{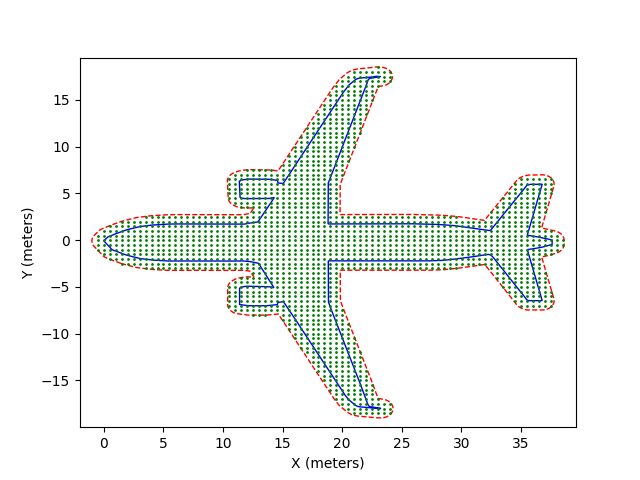}
    \caption{The discretisation of the area of interest (green dots).}
    \label{fig:a320-gridpoints}
\end{subfigure}
\caption{Different candidate modes of operation for the camera-based system.}
\label{fig:algorithm-3steps}
\end{figure}

Starting with the Airbus A320 aircraft as the base case, the initial step involves obtaining the top-view perimeter from the SVG file. This outline is then scaled to match a total length of 37.6 m, in accordance with the manufacturer's publicly available specifications. The result is depicted in Figure \ref{fig:a320-in-metres}.

The next step involves the extension of the perimeter to create a safe operating envelope to accommodate tolerances related to the operational scenario. For instance, in drone localisation, the value is 1 m to include a buffer space for the manoeuvres, but in defect detection, it is 0.5 m as there is no moving target and only aircraft misalignment should be considered. The result is illustrated in Figure \ref{fig:a320-inflated}.

Once the operating envelope is determined based on the specified target scenario, the next step involves discretising the region of interest. In cases where the focus is on the aircraft, such as in drone localisation, the area is transformed into a mosaic pattern based on the set discretisation distance, as demonstrated in Figure \ref{fig:a320-gridpoints}. Through experimentation, it was found that a value of 0.5 m maintains a reasonable balance between the accuracy of the solution and the processing time required. 

In the final stage, the algorithm, considering the discretised region of interest of the previous step and the suggested camera field of view projection, identifies the optimal camera layout for the bay. In the process, an overlap factor was also taken into account. This factor can affect system performance under varying conditions, especially with regard to transferring object localisation between cameras. If the scenario, such as defect localisation, does not necessitate it, the overlap is reduced. This parameter may also prove vital when stitching camera feeds into a larger image, as the presence of shared salient points becomes essential.

\subsubsection{Scenario A — Medium-Defect Mapping}
\begin{figure}
\centering
\begin{subfigure}[t]{0.48\textwidth}
    \includegraphics[width=\linewidth,valign=m]{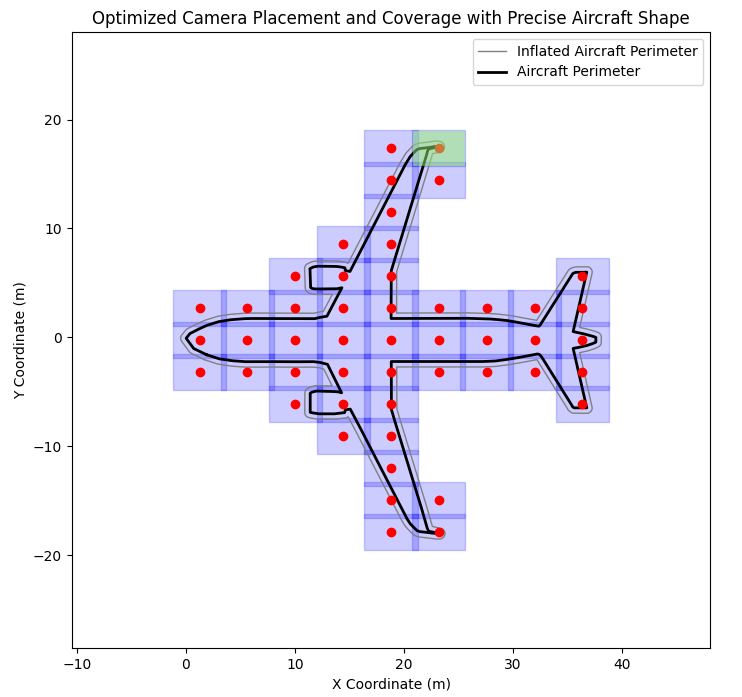}
    \caption{Optimal camera layout for the defect detection scenario (49 cameras). The target coverage area is 3.0 x 3.0 m, with a GSD of 0.89 mm/px.}
    \label{fig:dd_camera_based_layout}
\end{subfigure}
\hfill
\begin{subfigure}[t]{0.48\textwidth}
    \includegraphics[width=\linewidth,valign=m]{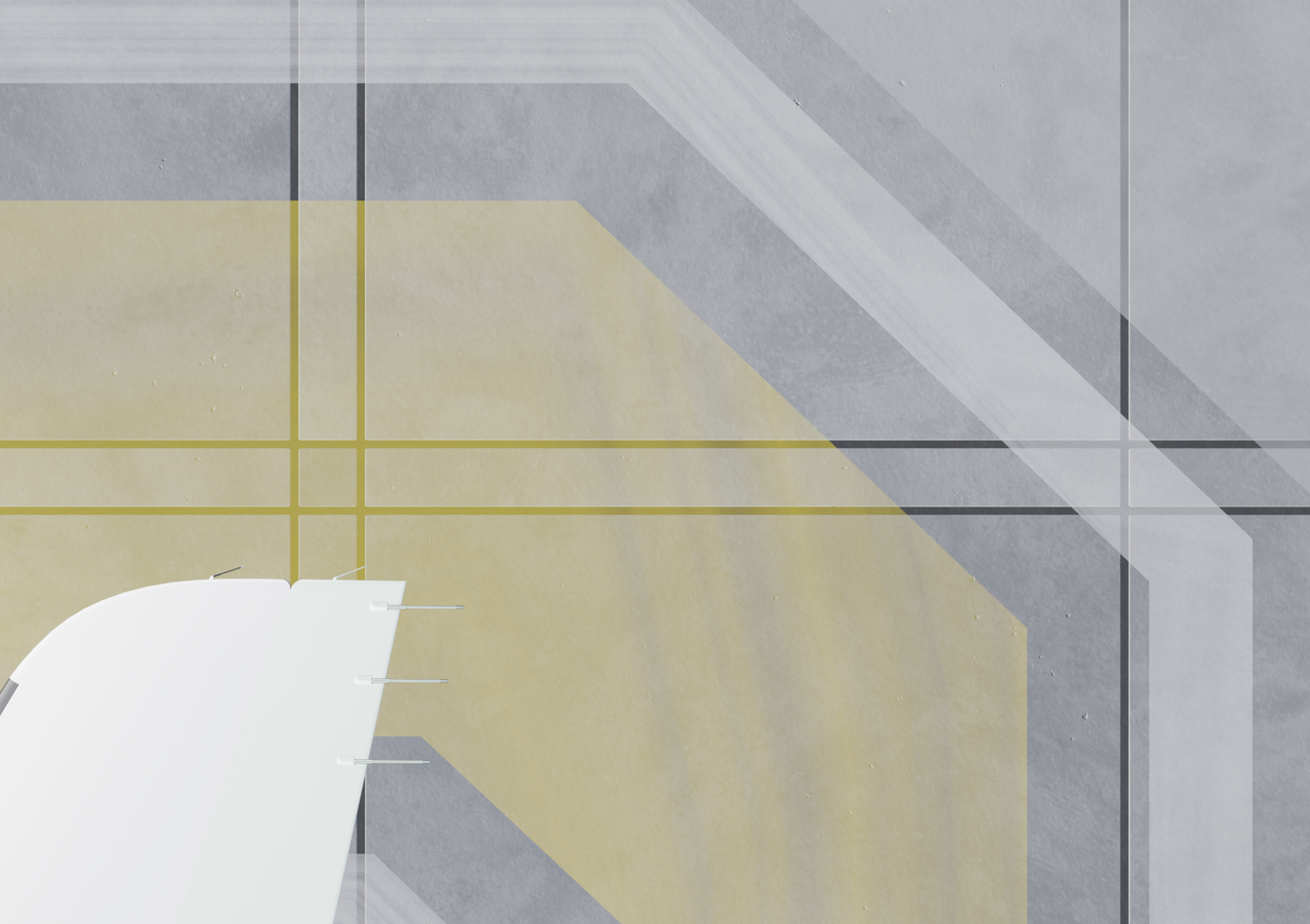}
    \caption{The view of the camera (Allied Vision 7) with the proposed lens (Techspec 7) in simulation targeting the end of the wing.}
    \label{fig:dd_camera_based_sim}
\end{subfigure}
\caption{Defect detection scenario for the camera-based system.}
\label{fig:dd_camera_based}
\end{figure}

The initial case of interest involves the detection of defects. This scenario focusses on identifying flaws present on the top surfaces of the aircraft, such as the fuselage and wings. The existing commercial systems, such as UWB and Mocap, are not designed for this scenario as they focus on localisation and tracking through the use of tags or reflective markers. This is notably intriguing because for maintenance technicians, it is safer to avoid working at heights with scaffolding and cherry pickers. During the experiments, it was realised that the only viable solution was the detection of medium-sized defects with an approximate size of 40 x 40 mm (Table \ref{tab:objects-or-interest-specs}). Detecting small defects requires sophisticated equipment, which is costly, and the number of cameras needed makes practical implementation unlikely. In medium-sized defects, the case is more realistic and affordable if the MRO industry decides to apply it. In this case, a potential defect will occupy approximately 45 px in the captured image. Given that the targets are located roughly between 16 and 19 meters away, corresponding to the aircraft's upper surface, the proposed configuration encompasses an area of 3 by 3 meters. To cover the entire surface of a narrow-body aircraft, 49 cameras are required (Figure \ref{fig:dd_camera_based}). Considering the additional cost of 3 PoE Ethernet switches, the cost adds up to a total of £76,809. In this cost estimate, cables are excluded since the installation cable length estimate was beyond the scope of the work. However, a representative price for 100 m of unshielded twisted pair (UTP) Category 6 cable is approximately £60. For this scenario, the cameras were configured to have a 10\% overlap in coverage to minimise the number of cameras needed. Moreover, as the target remains stationary, it is unnecessary to maintain a significant margin for the target transition between the cameras. Similarly, since the target is not moving, there is no negative impact that the suggested camera has a rolling shutter or the relatively low frame rate of 25 fps.

\subsubsection{Scenario B — Robots Localisation}
\begin{figure}
\centering
\begin{subfigure}[t]{0.48\textwidth}
    \includegraphics[width=\linewidth,valign=m]{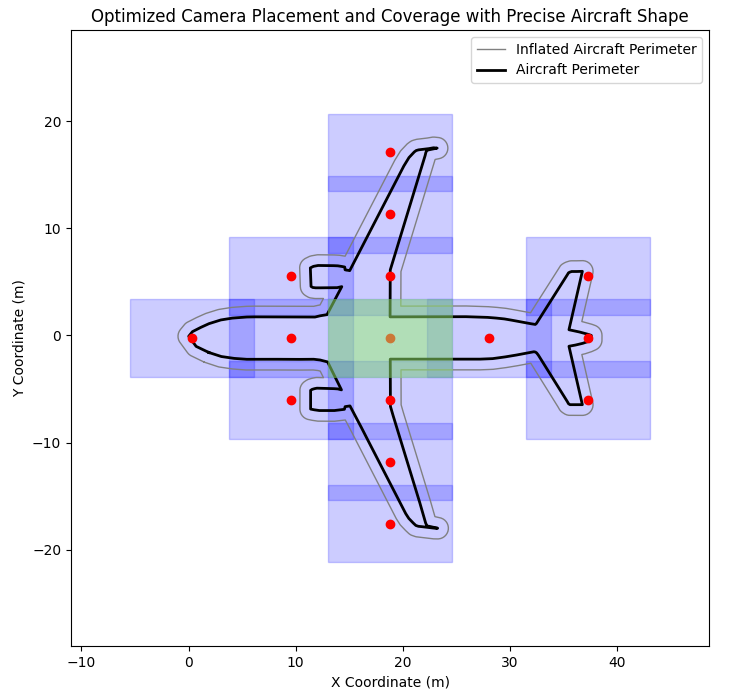}
    \caption{Optimal camera layout for the localisation scenario targeting drones (15 cameras). The target coverage area is 7.0 x 7.0 m, with a GSD of 6.02 mm/px.}
    \label{fig:ld_camera_based_layout}
\end{subfigure}
\hfill
\begin{subfigure}[t]{0.48\textwidth}
    \includegraphics[width=\linewidth,valign=m]{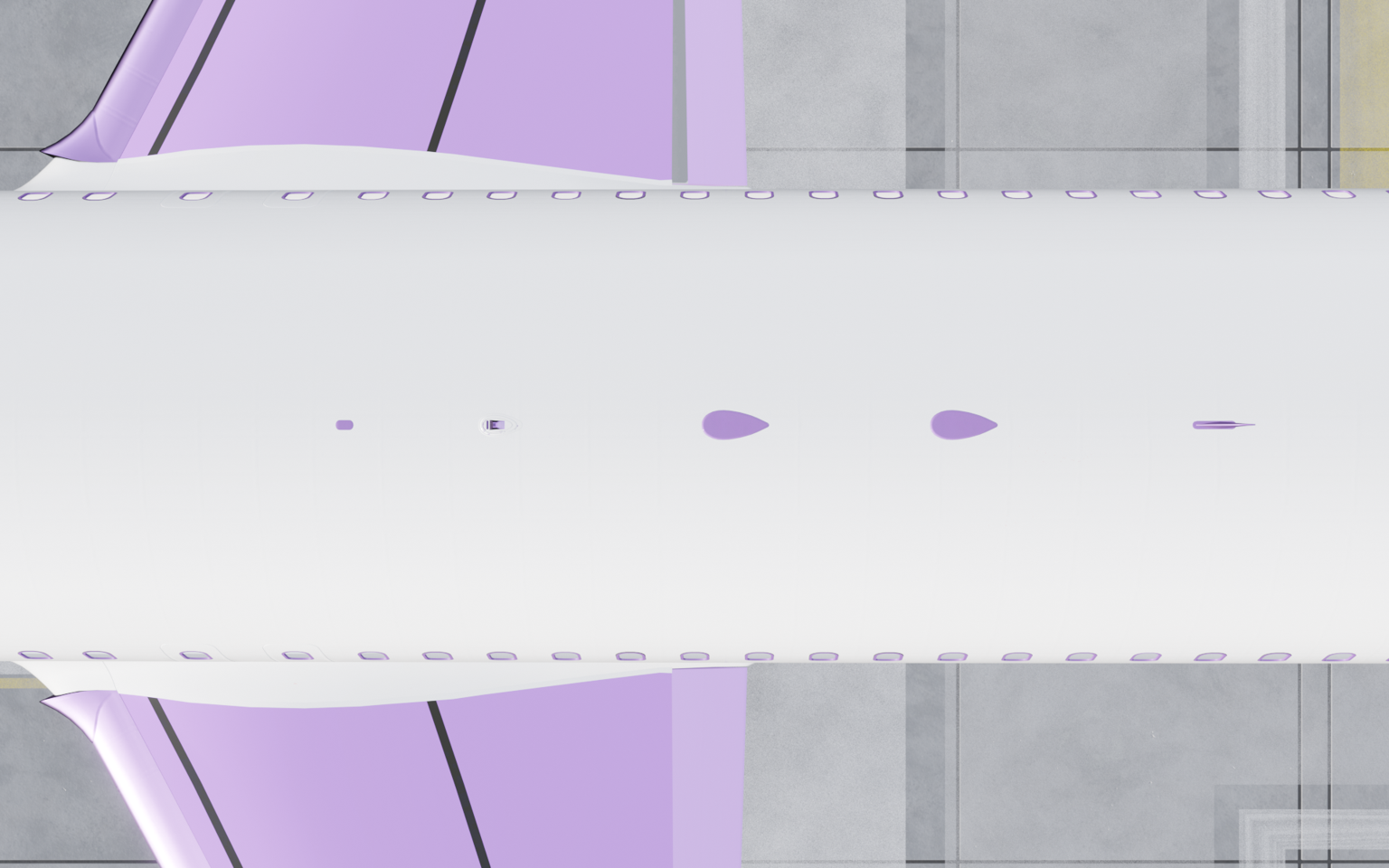}
    \caption{The view of the camera (Basler Ace 1) with the proposed lens (Techspec 4) in simulation targeting the centre of the fuselage.}
    \label{fig:ld_camera_based_sim}
\end{subfigure}
\caption{Localisation scenario targeting drones for the camera-based system.}
\label{fig:ld_camera_based}
\end{figure}

\begin{figure}
    \centering
    \includegraphics[width=0.75\linewidth]{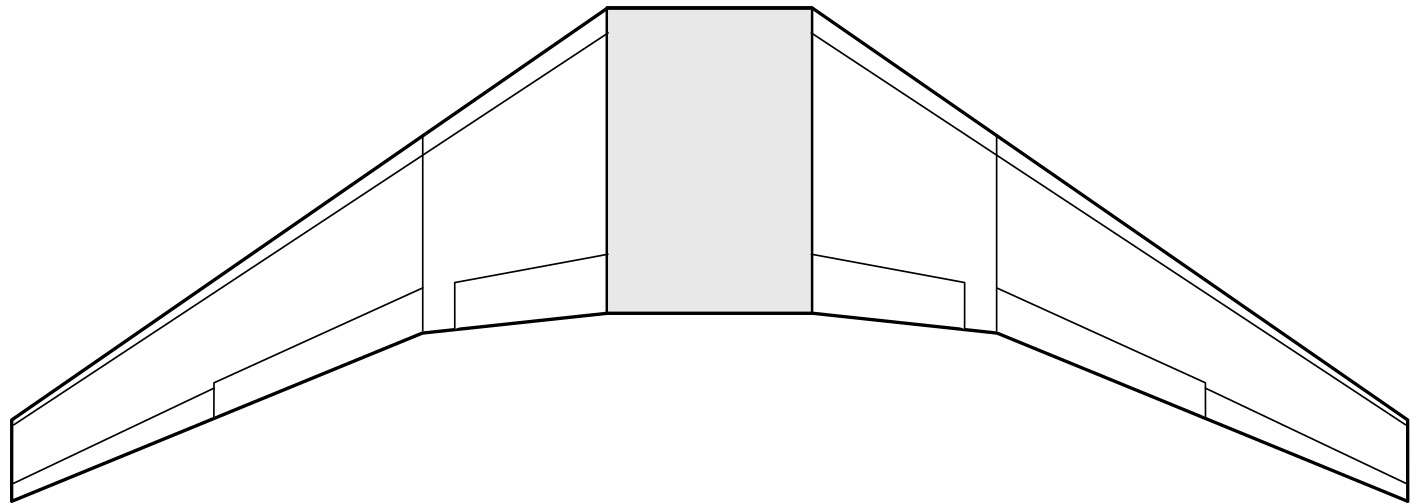}
    \caption{The Airbus A320 wings (Source online in \cite{AviationA320}). The area of both wings is 122.6 m\textsuperscript{2}}
    \label{fig:airbus_A320_wings}
\end{figure}

\begin{figure}
\centering
\begin{subfigure}[t]{0.48\textwidth}
    \includegraphics[width=\linewidth,valign=m]{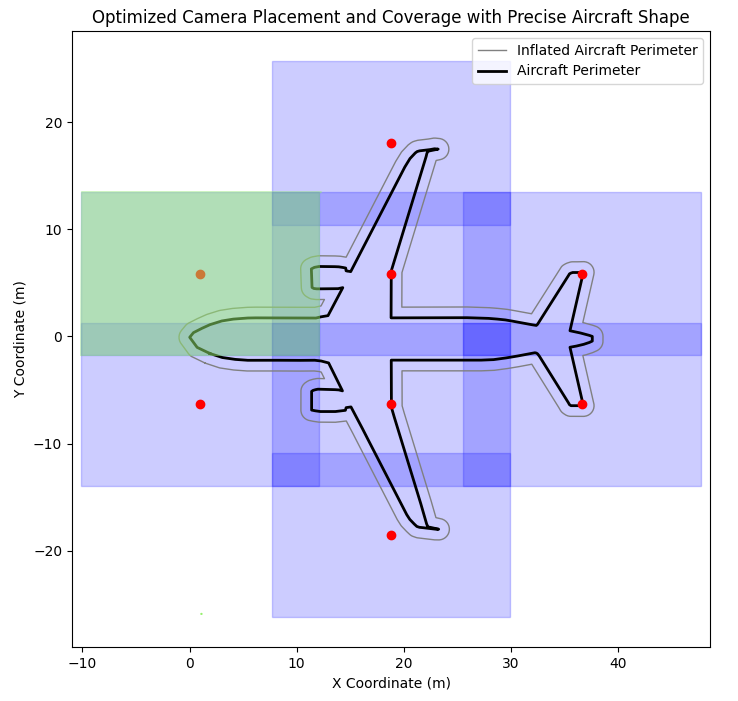}
    \caption{Optimal camera layout for the localisation targeting ground platforms (8 cameras). The target coverage area is 14.0 x 14.0 m, with a GSD of 6.94 mm/px.}
    \label{fig:lg_camera_based_layout}
\end{subfigure}
\hfill
\begin{subfigure}[t]{0.48\textwidth}
    \includegraphics[width=\linewidth,valign=m]{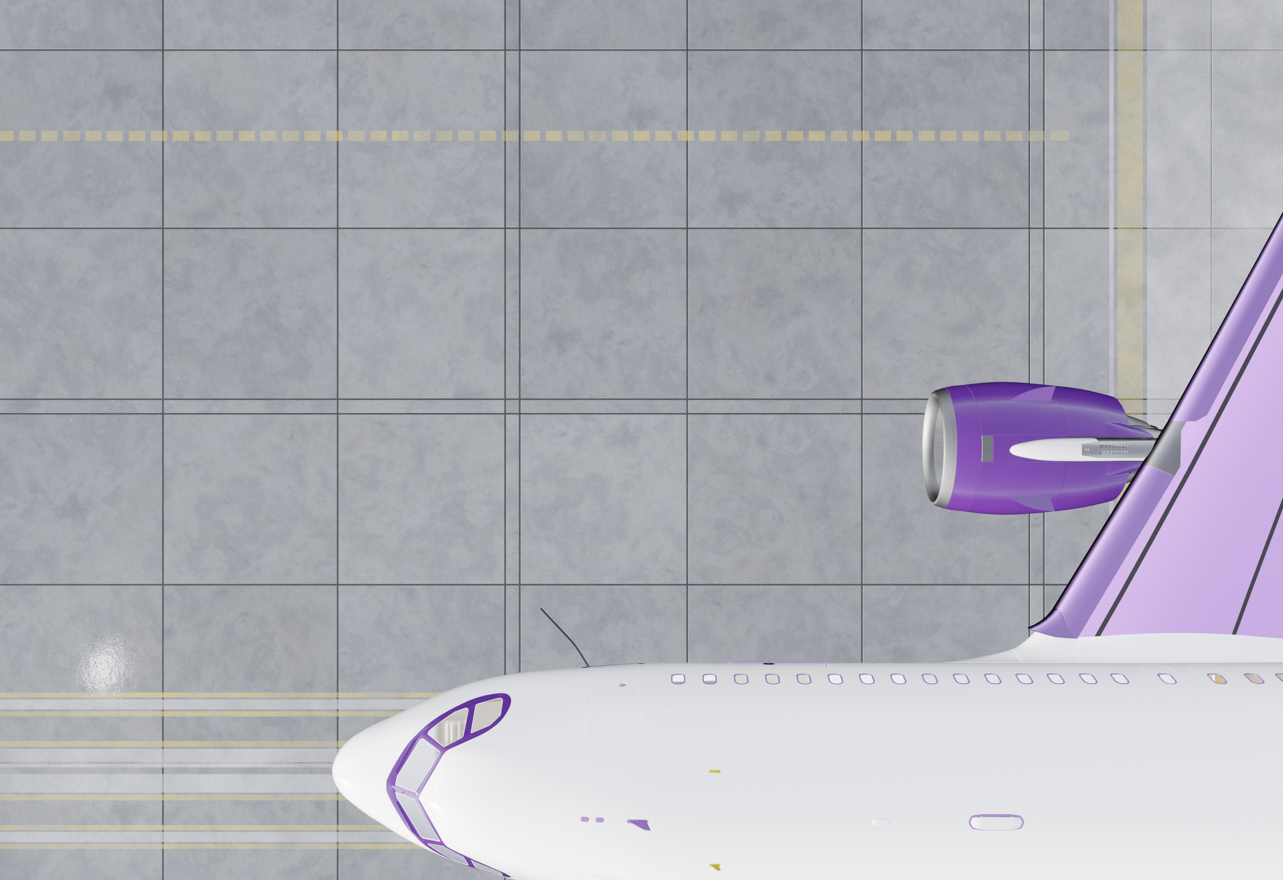}
    \caption{The view of the camera (Lucid 2) with the proposed lens (Techspec 3) in simulation targeting the centre of the fuselage.}
    \label{fig:lg_camera_based_sim}
\end{subfigure}
\caption{Localisation scenario targeting ground robotic platforms for the camera-based system.}
\label{fig:lg_camera_based}
\end{figure}

The localisation case has high importance since it can serve as an external system that provides the absolute pose of a robotic platform, aerial or ground, that can be fused with the onboard odometry and provide a reliable navigation. Based on the dimensions of the target and the distance from the camera (ceiling), there are two potential cases: the drones and the ground robotic platforms. Depending on the robot's operation space, the distance is approximately 15 to 18 m for the drones and 22 m for the ground platforms. The expected and allowable velocity of both robots is similar and can be satisfied from the frame rate of the cameras. The equations presented in \ref{eq:time_per_frame} can be employed to determine the distance travelled between frames. The camera for drones has 42 fps, and for the UGVs, 35 fps. With a maximum speed of 1.5 \si{m/s}, the distance between each frame for a 40 fps camera is 3.75 cm. In addition, both cameras suggested by the algorithm are equipped with a global shutter that is recommended for improved performance on moving targets. Figure \ref{fig:ld_camera_based} illustrates that covering the entire surface of a narrow-body aircraft requires 15 cameras. Incorporating a 24-port PoE Ethernet switch and UTP CAT6 cabling brings the total cost to around £16,500. In this scenario, an overlap factor of 20\% was established to facilitate the transfer of drone detection between neighbouring cameras. When focussing solely on the localisation of ground robotic platforms, the requirements for camera specifications related to object detection can be eased due to the larger size, leading to a configuration of eight cameras costing approximately £12,800 (Figure \ref{fig:lg_camera_based}). In this case, the overlap factor is adjusted to 20\%. In each case, it is crucial to emphasise that the chosen cameras feature global shutters, which are essential for preventing motion artefacts as the targets are moving robots.
\begin{equation}
    \label{eq:time_per_frame}
    \text { Time per frame }=\frac{1}{f}, \quad
    \text { Distance per frame }=v \times \text {Time per frame}
\end{equation}
Drones flying in proximity to the inspection surface can effectively address the challenges in defect detection scenarios where multiple cameras are typically needed, particularly limited to detecting medium- to large-sized defects. Instead of having 49 cameras and a budget of approximately £75,000, a single drone with a camera can identify small-sized defects of 10 mm with camera equipment that costs £2,061. In greater detail, running the camera specification algorithm, the Lucid camera 11 equipped with lens 3 can illustrate 10 mm defects at 45.45 px in the image. The assumption is that the drone flies approximately 1 m above the surface under inspection and that the field of view is limited to an area of 1x1 m. This operational setup generates images with appropriately-sized artefacts suitable for input into a DL algorithm that has been effectively trained to detect surface defects. One limitation of this method is the duration required for the drone to traverse and survey the entire area. To estimate the required time, consider the example of the A320's wing area. Each wing covers approximately 61.3 m\textsuperscript{2}, and combined, the total area for both wings is 122.6 m\textsuperscript{2} \cite{Airbus2024CustomerServices} (see Figure \ref{fig:airbus_A320_wings}). If the drone is photographing a 1 m\textsuperscript{2} patch, it needs approximately 61 images. The 122.6 mm\textsuperscript{2} figure refers to the flat 'shadow' area of the wing. In reality, the upper (and lower) skin of the wing is slightly curved (aerofoil shape), so the total area of the skin is somewhat larger, which can be approximated roughly to 63 m\textsuperscript{2}. To thoroughly survey the wing, the drone is generally flown in a "lawnmower" (boustrophedon) pattern \cite{Bahnemann2019RevisitingProblem}. This involves making straight, parallel flights separated by 1 m (the camera's swath), followed by a 180° turn at the end of each line to start the next one. It makes sense to fly each pass along the span (wing tip to wing root) because the A320 half-wing span is about 17.05 m, so each sweep is roughly 17 m long. In this case, covering the entire top of an A320 wing at 0.5 m/s, with a 1 m swath and 5 s per 180° turn, requires on the order of 2,5 minutes total. While this serves as a basic representation of the coverage pattern, it provides a ballpark estimate that requires additional refinement to accurately reflect the realistic time measurement.
The localisation scenario is interesting as it can be regarded as a direct competitor for commercially available systems, such as UWB and MoCap. Beyond the anticipated variations in performance resulting from the maturation of each method, there are also distinctions pertinent to the physics and the enabling technologies. The MoCap systems shine when it comes to accuracy, showing millimetre and sub-degree accuracy in the estimated position and heading. The rate of estimation is very high, in most cases, and is far above 100 Hz, depending on the cameras and system configuration. UWB systems offer centimetre-level accuracy, typically below 10 cm. However, estimating the heading is challenging, as the tag does not provide directional data. For orientation, it is advisable to employ two tags and determine their relative positions. The estimation frequency is quite high and varies with the tag, potentially reaching 30 Hz. In terms of estimating the position, both systems' performance depends on the line of sight between the target and the infrastructure. Both require at least 2 cameras/anchors to monitor the target. This is particularly useful for areas below the wing that still have some visibility from above. The selection depends on the available budget and the requirements of the localisation in terms of accuracy and speed.

\subsubsection{Scenario C — Ground Assets and Human Monitoring}
\begin{figure}
\centering
\begin{subfigure}{0.48\textwidth}
    \includegraphics[width=\linewidth]{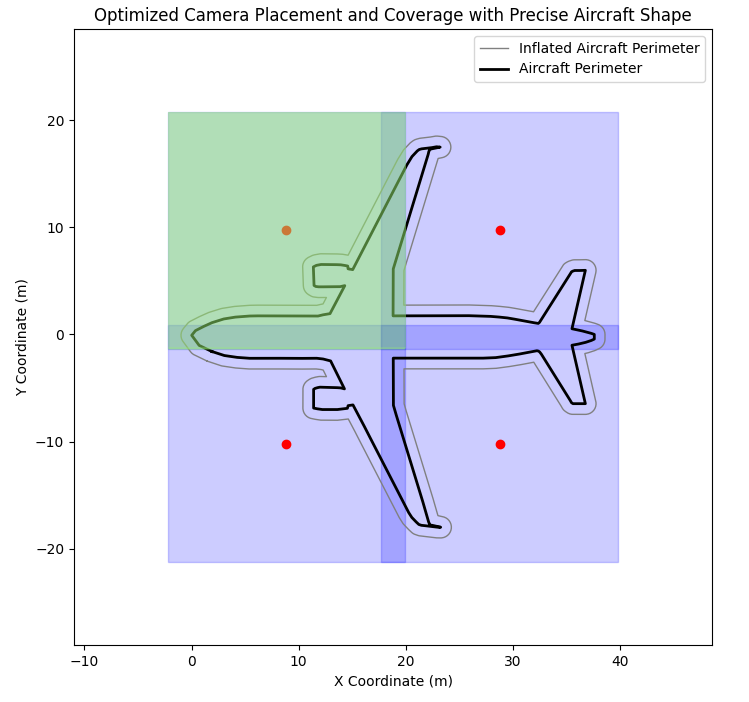}
    \caption{Optimal camera layout for the localisation targeting support vehicles (6 cameras). The target coverage area is 20.0 x 20.0 m, with a GSD of 4.91 mm/px.}
    \label{fig:mg_camera_based_layout}
\end{subfigure}
\hfill
\begin{subfigure}{0.48\textwidth}
    \includegraphics[width=\linewidth]{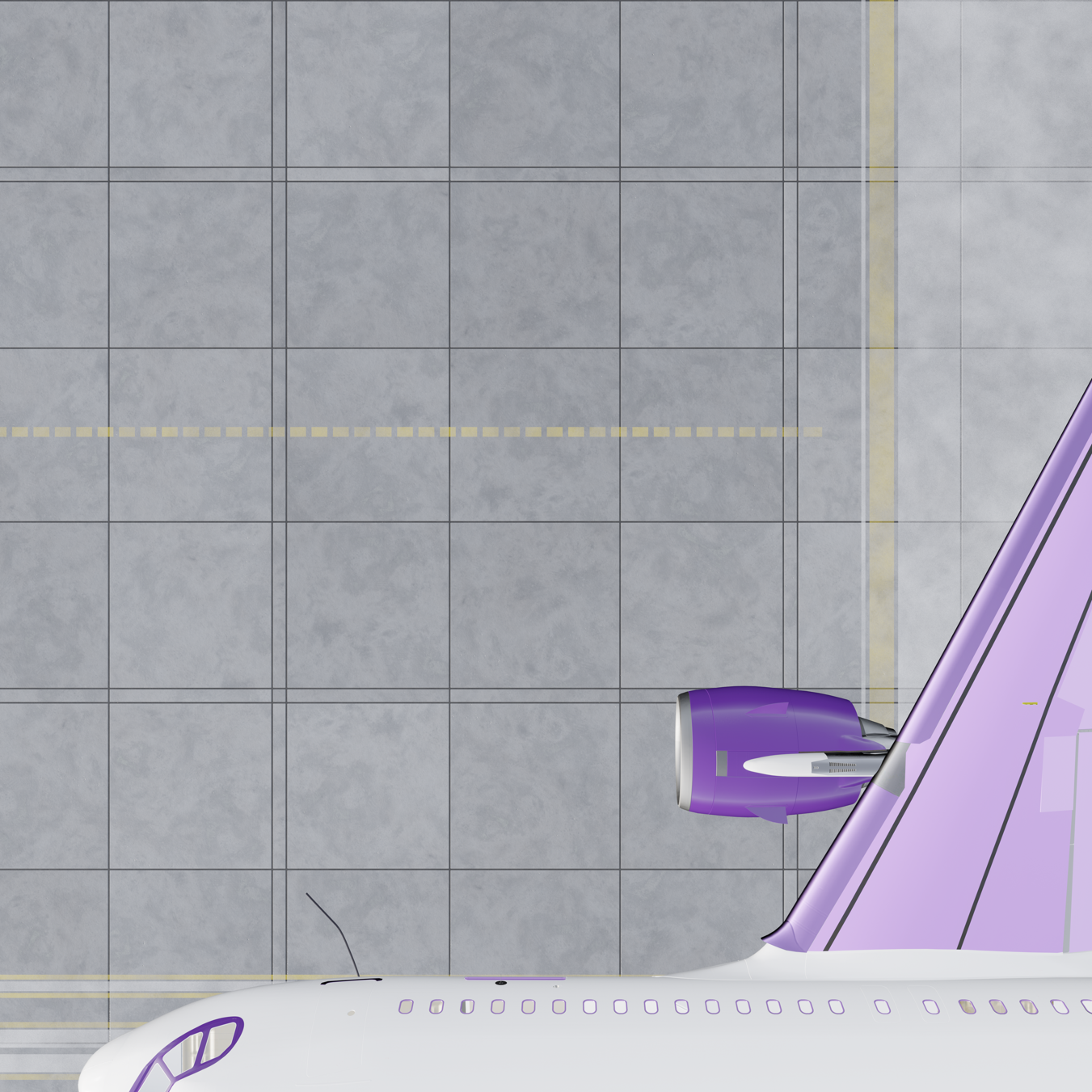}
    \caption{The view of the camera (Lucid 11) with the proposed lens (Techspec 3) in simulation targeting the front part of the fuselage.}
    \label{fig:mg_camera_based_sim}
\end{subfigure}
\caption{Monitoring scenario targeting ground support vehicles for the camera-based system.}
\label{fig:mg_camera_based}
\end{figure}

\begin{figure}
\centering
\begin{subfigure}[t]{0.48\textwidth}
    \includegraphics[width=\linewidth,valign=m]{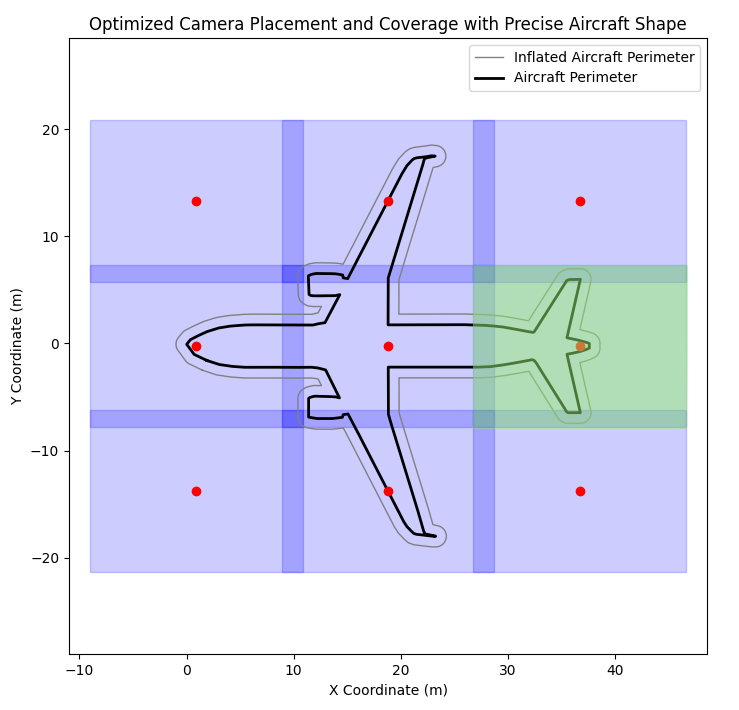}
    \caption{Optimal camera layout for the localisation targeting humans (9 cameras). The target coverage area is 15.0 x 15.0 m, with a GSD of 4.94 mm/px.}
    \label{fig:mh_camera_based_layout}
\end{subfigure}
\hfill
\begin{subfigure}[t]{0.48\textwidth}
    \includegraphics[width=\linewidth,valign=m]{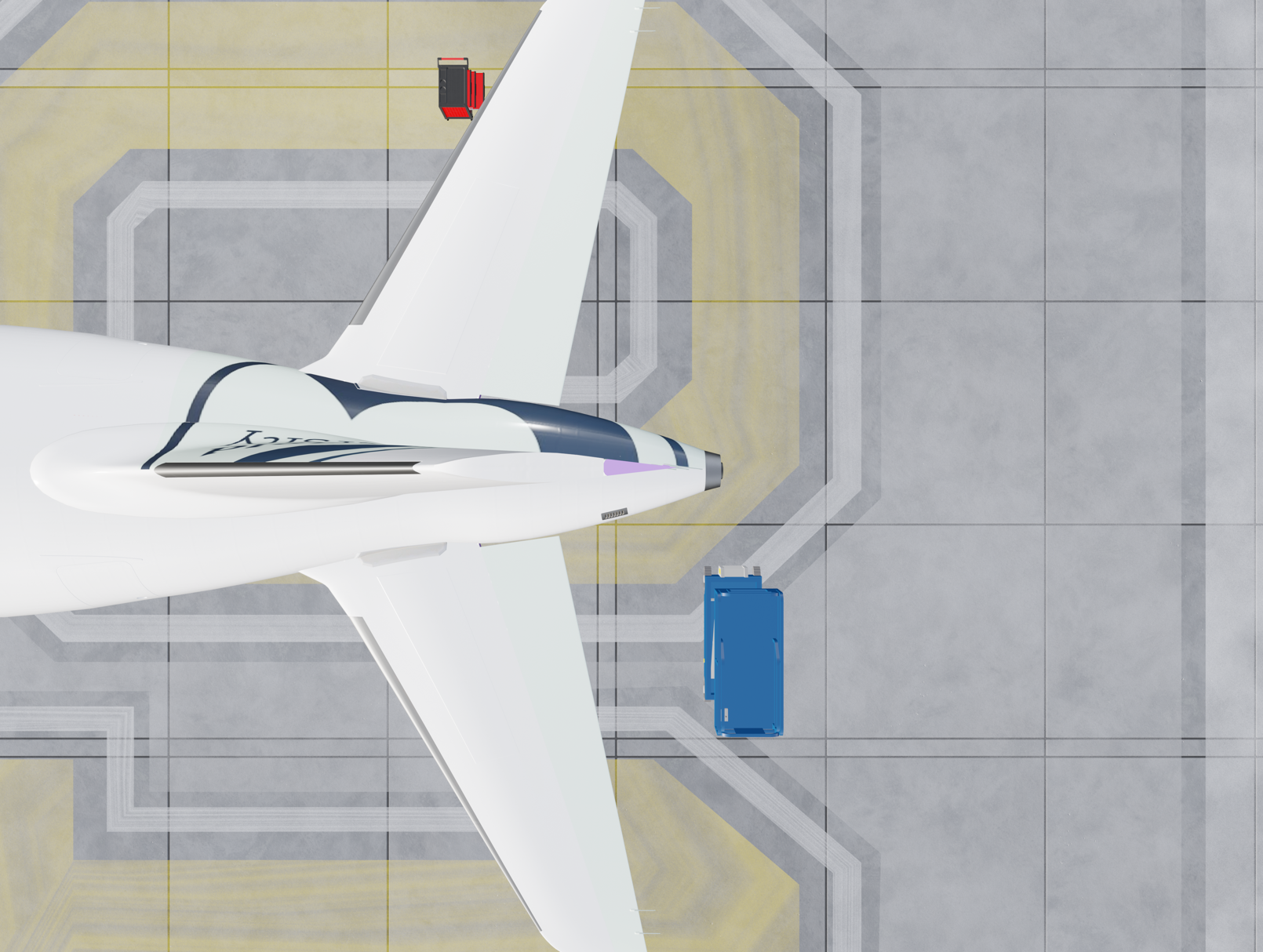}
    \caption{The view of the camera (Allied 3) with the proposed lens (Techspec 2) in simulation targeting the front part of the fuselage.}
    \label{fig:mh_camera_based_sim}
\end{subfigure}
\caption{Monitoring scenario including humans for the camera-based system.}
\label{fig:mh_camera_based}
\end{figure}

The monitoring case focuses on tracking moving assets on the ground, including ground support vehicles, personnel, tool cribs, ground power units (GPUs) and tugs. Typically, height assumptions for viewing depth range from 1 to 2 meters, resulting in an average distance of 21.5 meters from the ceiling. From an overhead perspective, there is a significant difference in the area occupied by a vehicle compared to that of a human (Table \ref{tab:objects-or-interest-specs}). This is why camera specifications are designed separately for the two distinct scenarios. The allowed maximum speed for terrestrial vehicles is 2 \si{\metre\per\second}, while the typical average walking speed for a person is about 1.27 \si{\metre\per\second}. In the first case, the chosen camera records 5 cm per frame, while in the second case, it captures 9.4 cm per frame. For ground vehicles, six cameras are necessary (Figure \ref{fig:mg_camera_based}); however, if the targets include humans, the requirement increases to nine cameras (Figure \ref{fig:mh_camera_based}). The price, which includes a 24-port PoE Ethernet Switch and UTP CAT6 cables, is approximately £8,800 and £17,200, respectively.

\subsubsection{Comparative blueprint synthesis}

To consolidate the design-to-cost blueprints discussed, Table \ref{tab:blueprints} contrasts the five ceiling-camera scenarios with the MoCap and UWB reference systems.  The snapshot highlights how the intended application, bill-of-materials footprint, and technology maturity interact with overall budget, providing a concise decision aid for hangar planners.

\begin{longtable}{@{}c p{2.5cm} p{3.5cm} c c@{}}
    \caption{Comparison of localisation and monitoring blueprints proposed.}
    \label{tab:blueprints} \\

    \toprule
    \textbf{Blueprint} & 
    \textbf{Application} &
    \textbf{Indicative on--bay equipment} & 
    \textbf{Cost (£k)} & 
    \textbf{Maturity} \\
    \midrule
    \endfirsthead

    \toprule
    \textbf{Blueprint} & 
    \textbf{Application} &
    \textbf{Indicative on--bay equipment} & 
    \textbf{Cost (£k)} & 
    \textbf{Maturity} \\
    \midrule
    \endhead

    \midrule
    \multicolumn{5}{r}{\small\em Continued on next page} \\
    \midrule
    \endfoot

    \bottomrule
    \endlastfoot
    
    Vision - A & Medium-sized defect detection on upper surfaces & 49x20MP rolling shutter GigE cameras + 3 PoE switches & 76.8 & Low TRL\\ 
    Vision - B1 & Drone localisation & 15x2.3MP global shutter GigE cameras & 16.5 & Low TRL \\ 
    Vision - B2 & Ground robot localisation & 8x7.1MP global shutter GigE cameras & 12.8 & Low TRL\\ 
    Vision - C1 & Vehicle tracking  & 6x12.3MP global shutter GigE cameras & 8.8 & Low TRL\\ 
    Vision - C2 & Personnel tracking & 9x12.4MP global shutter GigE cameras & 17.2 & Low TRL\\ 
    UWB & Asset localisation / tracking & 10--25 anchors, tags, timing units, PoE switches & 49.0 & High TRL\\ 
    MoCap & High-precision localisation & 12 IR cameras, active markers, sync hub & 190.0 & High TRL\\ 
\end{longtable}

\section{Conclusions}
This study presents the first comprehensive techno-economic roadmap for localisation, artefact detection, and monitoring within a smart hangar. Rigorously benchmarking MoCap, UWB and camera-based systems across three representative scenarios -robot location, asset tracking, and surface defect detection - enables the demonstration of the interactions between accuracy, coverage, and cost within a 40 × 50 m bay. The proposed dual-layer framework combines market-driven camera lens selection with optimisation algorithms, demonstrating how a vision system can be deployed in various operational scenarios. Design-to-cost case studies demonstrate that only 15 global-shutter GigE cameras are sufficient for drone localisation, 9 for monitoring, whereas 49 high-resolution units enable complete airframe midsize defect mapping.

These findings have immediate practical value for MRO planners. The blueprint table furnishes a ready-made bill-of-materials envelope, reducing specification time, and derisking procurement. Future work should address two open challenges. First, large-scale certification will require extended endurance trials to characterise failure modes under dense reflections and dynamic occlusions. Second, development and validation of the customised deep-learning approaches to support the camera-based solution back-end. Resolving these issues will propel the transition from isolated demonstrators to fully integrated 'Smart Hangars' in which sensing infrastructures, robots, and human technicians collaborate seamlessly to enhance safety, turnaround time, and sustainability.

\section*{ACKNOWLEDGMENTS}
We extend our gratitude to Martin Lewis, Andy Ward, and Dami Phillips for their invaluable support and insights on commercially available systems.

\clearpage


\printbibliography

@article{Bentham2016AHeight,
    title = {{A century of trends in adult human height}},
    year = {2016},
    journal = {eLife},
    author = {Bentham, James and Di Cesare, Mariachiara and Stevens, Gretchen A. and Zhou, Bin and Bixby, Honor and Cowan, Melanie and Fortunato, Léa and Bennett, James E. and Danaei, Goodarz and Hajifathalian, Kaveh and Lu, Yuan and Riley, Leanne M. and Laxmaiah, Avula and Kontis, Vasilis and Paciorek, Christopher J. and Riboli, Elio and Ezzati, Majid and Abdeen, Ziad A. and Hamid, Zargar Abdul and Abu-Rmeileh, Niveen M. and Acosta-Cazares, Benjamin and Adams, Robert and Aekplakorn, Wichai and Aguilar-Salinas, Carlos A. and Agyemang, Charles and Ahmadvand, Alireza and Ahrens, Wolfgang and Alhazzaa, Hazzaa M. and Al-Othman, Amani Rashed and Raddadi, Rajaa Al and Ali, Mohamed M. and Alkerwi, Ala’A and Alvarez-Pedrerol, Mar and Aly, Eman and Amouyel, Philippe and Amuzu, Antoinette and Andersen, Lars Bo and Anderssen, Sigmund A. and Anjana, Ranjit Mohan and Aounallah-Skhiri, Hajer and Ariansen, Inger and Aris, Tahir and Arlappa, Nimmathota and Arveiler, Dominique and Assah, Felix K. and Avdicov{\'{a}}, Mária and Azizi, Fereidoun and Babu, Bontha V. and Bahijri, Suhad and Balakrishna, Nagalla and Bandosz, Piotr and Banegas, José R. and Barbagallo, Carlo M. and Barcel{\'{o}}, Alberto and Barkat, Amina and Barros, Mauro V. and Bata, Iqbal and Batieha, Anwar M. and Batista, Rosangela L. and Baur, Louise A. and Beaglehole, Robert and Romdhane, Habiba Ben and Benet, Mikhail and Bernabe-Ortiz, Antonio and Bernotiene, Gailute and Bettiol, Heloisa and Bhagyalaxmi, Aroor and Bharadwaj, Sumit and Bhargava, Santosh K. and Bhatti, Zaid and Bhutta, Zulfiqar A. and Bi, Hongsheng and Bi, Yufang and Bjerregaard, Peter and Bjertness, Espen and Bjertness, Marius B. and Bj{\"{o}}rkelund, Cecilia and Blokstra, Anneke and Bo, Simona and Bobak, Martin and Boddy, Lynne M. and Boehm, Bernhard O. and Boeing, Heiner and Boissonnet, Carlos P. and Bongard, Vanina and Bovet, Pascal and Braeckman, Lutgart and Bragt, Marjolijn C.E. and Brajkovich, Imperia and Branca, Francesco and Breckenkamp, Juergen and Brenner, Hermann and Brewster, Lizzy M. and Brian, Garry R. and Bruno, Graziella and Bueno-De-mesquita, H. B. and Bugge, Anna and Burns, Con and de Le{\'{o}}n, Antonio Cabrera and Cacciottolo, Joseph and Cama, Tilema and Cameron, Christine and Camolas, José and Can, Günay and C{\^{a}}ndido, Ana Paula C. and Capuano, Vincenzo and Cardoso, Viviane C. and Carlsson, Axel C. and Carvalho, Maria J. and Casanueva, Felipe F. and Casas, Juan Pablo and Caserta, Carmelo A. and Chamukuttan, Snehalatha and Chan, Angelique W. and Chan, Queenie and Chaturvedi, Himanshu K. and Chaturvedi, Nishi and Chen, Chien Jen and Chen, Fangfang and Chen, Huashuai and Chen, Shuohua and Chen, Zhengming and Cheng, Ching Yu and Chetrit, Angela and Chiolero, Arnaud and Chiou, Shu Ti and Chirita-Emandi, Adela and Cho, Belong and Cho, Yumi and Christensen, Kaare and Chudek, Jerzy and Cifkova, Renata and Claessens, Frank and Clays, Els and Concin, Hans and Cooper, Cyrus and Cooper, Rachel and Coppinger, Tara C. and Costanzo, Simona and Cottel, Dominique and Cowell, Chris and Craig, Cora L. and Crujeiras, Ana B. and D’Arrigo, Graziella and D’Orsi, Eleonora and Dallongeville, Jean and Damasceno, Albertino and Damsgaard, Camilla T. and Dankner, Rachel and Dauchet, Luc and De Backer, Guy and De Bacquer, Dirk and de Gaetano, Giovanni and De Henauw, Stefaan and De Smedt, Delphine and Deepa, Mohan and Deev, Alexander D. and Dehghan, Abbas and Delisle, Hélène and Delpeuch, Francis and Deschamps, Valérie and Dhana, Klodian and Di Castelnuovo, Augusto F. and Dias-Da-costa, Juvenal Soares and Diaz, Alejandro and Djalalinia, Shirin and Do, Ha T.P. and Dobson, Annette J. and Donfrancesco, Chiara and Donoso, Silvana P. and D{\"{o}}ring, Angela and Doua, Kouamelan and Drygas, Wojciech and Dzerve, Vilnis and Egbagbe, Eruke E. and Eggertsen, Robert and Ekelund, Ulf and El Ati, Jalila and Elliott, Paul and Engle-Stone, Reina and Erasmus, Rajiv T. and Erem, Cihangir and Eriksen, Louise and Escobedo-De la Pe{\~{n}}a, Jorge and Evans, Alun and Faeh, David and Fall, Caroline H. and Farzadfar, Farshad and Felix-Redondo, Francisco J. and Ferguson, Trevor S. and Fern{\'{a}}ndez-Berg{\'{e}}s, Daniel and Ferrante, Daniel and Ferrari, Marika and Ferreccio, Catterina and Ferrieres, Jean and Finn, Joseph D. and Fischer, Krista and Flores, Eric Monterubio and F{\"{o}}ger, Bernhard and Foo, Leng Huat and Forslund, Ann Sofie and Forsner, Maria and Fortmann, Stephen P. and Fouad, Heba M. and Francis, Damian K. and Franco, Maria Do Carmo and Franco, Oscar H. and Frontera, Guillermo and Fuchs, Flavio D. and Fuchs, Sandra C. and Fujita, Yuki and Furusawa, Takuro and Gaciong, Zbigniew and Gafencu, Mihai and Gareta, Dickman and Garnett, Sarah P. and Gaspoz, Jean Michel and Gasull, Magda and Gates, Louise and Geleijnse, Johanna M. and Ghasemian, Anoosheh and Giampaoli, Simona and Gianfagna, Francesco and Giovannelli, Jonathan and Giwercman, Aleksander and Goldsmith, Rebecca A. and Gon{\c{c}}alves, Helen and Gross, Marcela Gonzalez and Rivas, Juan P.González and Gorbea, Mariano Bonet and Gottrand, Frederic and Graff-Iversen, Sidsel and Grafnetter, Dušan and Grajda, Aneta and Grammatikopoulou, Maria G. and Gregor, Ronald D. and Grodzicki, Tomasz and Gr{\o}ntved, Anders and Gruden, Grabriella and Grujic, Vera and Gu, Dongfeng and Gualdi-Russo, Emanuela and Guan, Ong Peng and Gudnason, Vilmundur and Guerrero, Ramiro and Guessous, Idris and Guimaraes, Andre L. and Gulliford, Martin C. and Gunnlaugsdottir, Johanna and Gunter, Marc and Guo, Xiuhua and Guo, Yin and Gupta, Prakash C. and Gureje, Oye and Gurzkowska, Beata and Gutierrez, Laura and Gutzwiller, Felix and Halkj{\ae}r, Jytte and Hambleton, Ian R. and Hardy, Rebecca and Kumar, Rachakulla Hari and Hata, Jun and Hayes, Alison J. and He, Jiang and Hendriks, Marleen Elisabeth and Cadena, Leticia Hernandez and Herrala, Sauli and Heshmat, Ramin and Hihtaniemi, Ilpo Tapani and Ho, Sai Yin and Ho, Suzanne C. and Hobbs, Michael and Hofman, Albert and Hormiga, Claudia M. and Horta, Bernardo L. and Houti, Leila and Howitt, Christina and Htay, Thein Thein and Htet, Aung Soe and Htike, Maung Maung Than and Hu, Yonghua and Husseini, Abdullatif and Huu, Chinh Nguyen and Huybrechts, Inge and Hwalla, Nahla and Iacoviello, Licia and Iannone, Anna G. and Ibrahim, Mohsen M. and Ikeda, Nayu and Ikram, M. Arfan and Irazola, Vilma E. and Islam, Muhammad and Ivkovic, Vanja and Iwasaki, Masanori and Jackson, Rod T. and Jacobs, Jeremy M. and Jafar, Tazeen and Jamil, Kazi M. and Jamrozik, Konrad and Janszky, Imre and Jasienska, Grazyna and Jelakovic, Bojan and Jiang, Chao Qiang and Joffres, Michel and Johansson, Mattias and Jonas, Jost B. and J{\o}rgensen, Torben and Joshi, Pradeep and Juolevi, Anne and Jurak, Gregor and Jure{\v{s}}a, Vesna and Kaaks, Rudolf and Kafatos, Anthony and Kalter-Leibovici, Ofra and Kapantais, Efthymios and Kasaeian, Amir and Katz, Joanne and Kaur, Prabhdeep and Kavousi, Maryam and Keil, Ulrich and Boker, Lital Keinan and Kein{\"{a}}nen-Kiukaanniemi, Sirkka and Kelishadi, Roya and Kemper, Han C.G. and Kengne, Andre P. and Kersting, Mathilde and Key, Timothy and Khader, Yousef Saleh and Khalili, Davood and Khang, Young Ho and Khaw, Kay Tee H. and Khouw, Ilse M.S.L. and Kiechl, Stefan and Killewo, Japhet and Kim, Jeongseon and Klimont, Jeannette and Klumbiene, Jurate and Koirala, Bhawesh and Kolle, Elin and Kolsteren, Patrick and Korrovits, Paul and Koskinen, Seppo and Kouda, Katsuyasu and Koziel, Slawomir and Kratzer, Wolfgang and Krokstad, Steinar and Kromhout, Daan and Kruger, Herculina S. and Kubinova, Ruzena and Kujala, Urho M. and Kula, Krzysztof and Kulaga, Zbigniew and Kumar, R. Krishna and Kurjata, Pawel and Kusuma, Yadlapalli S. and Kuulasmaa, Kari and Kyobutungi, Catherine and Laamiri, Fatima Zahra and Laatikainen, Tiina and Lachat, Carl and Laid, Youcef and Lam, Tai Hing and Landrove, Orlando and Lanska, Vera and Lappas, Georg and Larijani, Bagher and Laugsand, Lars E. and Bao, Khanh Le Nguyen and Le, Tuyen D. and Leclercq, Catherine and Lee, Jeannette and Lee, Jeonghee and Lehtim{\"{a}}ki, Terho and Lekhraj, Rampal and Le{\'{o}}n-Mu{\~{n}}oz, Luz M. and Li, Yanping and Lilly, Christa L. and Lim, Wei Yen and Lima-Costa, M. Fernanda and Lin, Hsien Ho and Lin, Xu and Linneberg, Allan and Lissner, Lauren and Litwin, Mieczyslaw and Liu, Jing and Lorbeer, Roberto and Lotufo, Paulo A. and Lozano, José Eugenio and Luksiene, Dalia and Lundqvist, Annamari and Lunet, Nuno and Lytsy, Per and Ma, Guansheng and Ma, Jun and Machado-Coelho, George L.L. and Machi, Suka and Maggi, Stefania and Magliano, Dianna J. and Maire, Bernard and Makdisse, Marcia and Malekzadeh, Reza and Malhotra, Rahul and Rao, Kodavanti Mallikharjuna and Malyutina, Sofia and Manios, Yannis and Mann, Jim I. and Manzato, Enzo and Margozzini, Paula and Markey, Oonagh and Marques-Vidal, Pedro and Marrugat, Jaume and Martin-Prevel, Yves and Martorell, Reynaldo and Masoodi, Shariq R. and Mathiesen, Ellisiv B. and Matsha, Tandi E. and Mazur, Artur and Mbanya, Jean Claude N. and McFarlane, Shelly R. and McGarvey, Stephen T. and McKee, Martin and McLachlan, Stela and McLean, Rachael M. and McNulty, Breige A. and Yusof, Safiah Md and Mediene-Benchekor, Sounnia and Meirhaeghe, Aline and Meisinger, Christa and Menezes, Ana Maria B. and Mensink, Gert B.M. and Meshram, Indrapal I. and Metspalu, Andres and Mi, Jie and Michaelsen, Kim F. and Mikkel, Kairit and Miller, Jody C. and Miquel, Juan Francisco and Miranda, J. Jaime and Mi{\v{s}}igoj-Durakovic, Marjeta and Mohamed, Mostafa K. and Mohammad, Kazem and Mohammadifard, Noushin and Mohan, Viswanathan and Yusoff, Muhammad Fadhli Mohd and Molbo, Drude and M{\o}ller, Niels C. and Moln{\'{a}}r, Dénes and Mondo, Charles K. and Monterrubio, Eric A. and Monyeki, Kotsedi Daniel K. and Moreira, Leila B. and Morejon, Alain and Moreno, Luis A. and Morgan, Karen and Mortensen, Erik Lykke and Moschonis, George and Mossakowska, Malgorzata and Mostafa, Aya and Mota, Jorge and Motlagh, Mohammad Esmaeel and Motta, Jorge and Mu, Thet Thet and Muiesan, Maria Lorenza and M{\"{u}}ller-Nurasyid, Martina and Murphy, Neil and Mursu, Jaakko and Murtagh, Elaine M. and Musa, Kamarul Imran and Musil, Vera and Nagel, Gabriele and Nakamura, Harunobu and N{\'{a}}me{\v{s}}n{\'{a}}, Jana and Nang, Ei Ei K. and Nangia, Vinay B. and Nankap, Martin and Narake, Sameer and Navarrete-Mu{\~{n}}oz, Eva Maria and Neal, William A. and Nenko, Ilona and Neovius, Martin and Nervi, Flavio and Neuhauser, Hannelore K. and Nguyen, Nguyen D. and Nguyen, Quang Ngoc and Nieto-Mart{\'{i}}nez, Ramfis E. and Ning, Guang and Ninomiya, Toshiharu and Nishtar, Sania and Noale, Marianna and Norat, Teresa and Noto, Davide and Nsour, Mohannad Al and O’Reilly, Dermot and Oh, Kyungwon and Olayan, Iman H. and Olinto, Maria Teresa Anselmo and Oltarzewski, Maciej and Omar, Mohd A. and Onat, Altan and Ordunez, Pedro and Ortiz, Ana P. and Osler, Merete and Osmond, Clive and Ostojic, Sergej M. and Otero, Johanna A. and Overvad, Kim and Owusu-Dabo, Ellis and Paccaud, Fred Michel and Padez, Cristina and Pahomova, Elena and Pajak, Andrzej and Palli, Domenico and Palloni, Alberto and Palmieri, Luigi and Panda-Jonas, Songhomitra and Panza, Francesco and Parnell, Winsome R. and Parsaeian, Mahboubeh and Pecin, Ivan and Pednekar, Mangesh S. and Peeters, Petra H. and Peixoto, Sergio Viana and Peltonen, Markku and Pereira, Alexandre C. and P{\'{e}}rez, Cynthia M. and Peters, Annette and Petkeviciene, Janina and Peykari, Niloofar and Pham, Son Thai and Pigeot, Iris and Pikhart, Hynek and Pilav, Aida and Pilotto, Lorenza and Pistelli, Francesco and Pitakaka, Freda and Piwonska, Aleksandra and Plans-Rubi{\'{o}}, Pedro and Poh, Bee Koon and Porta, Miquel and Portegies, Marileen L.P. and Poulimeneas, Dimitrios and Pradeepa, Rajendra and Prashant, Mathur and Price, Jacqueline F. and Puiu, Maria and Punab, Margus and Qasrawi, Radwan F. and Qorbani, Mostafa and Bao, Tran Quoc and Radic, Ivana and Radisauskas, Ricardas and Rahman, Mahmudur and Raitakari, Olli and Raj, Manu and Rao, Sudha Ramachandra and Ramachandran, Ambady and Ramke, Jacqueline and Ramos, Rafel and Rampal, Sanjay and Rasmussen, Finn and Redon, Josep and Reganit, Paul Ferdinand M. and Ribeiro, Robespierre and Rigo, Fernando and de Wit, Tobias F.Rinke and Ritti-Dias, Raphael M. and Rivera, Juan A. and Robinson, Sian M. and Robitaille, Cynthia and Rodr{\'{i}}guez-Artalejo, Fernando and Rodriguez-Perez, María Del Cristo and Rodr{\'{i}}guez-Villamizar, Laura A. and Rojas-Martinez, Rosalba and Rojroongwasinkul, Nipa and Romaguera, Dora and Ronkainen, Kimmo and Rosengren, Annika and Rouse, Ian and Rubinstein, Adolfo and R{\"{u}}hli, Frank J. and Rui, Ornelas and Ruiz-Betancourt, Blanca Sandra and Horimoto, Andrea R.V.Russo and Rutkowski, Marcin and Sabanayagam, Charumathi and Sachdev, Harshpal S. and Saidi, Olfa and Salanave, Benoit and Martinez, Eduardo Salazar and Salomaa, Veikko and Salonen, Jukka T. and Salvetti, Massimo and S{\'{a}}nchez-Abanto, Jose and {Sandjaja} and Sans, Susana and Santos, Diana A. and Santos, Osvaldo and Dos Santos, Renata Nunes and Santos, Rute and Saramies, Jouko L. and Sardinha, Luis B. and Sarrafzadegan, Nizal and Saum, Kai Uwe and Savva, Savvas C. and Scazufca, Marcia and Rosario, Angelika Schaffrath and Schargrodsky, Herman and Schienkiewitz, Anja and Schmidt, Ida Maria and Schneider, Ione J. and Schultsz, Constance and Schutte, Aletta E. and Sein, Aye Aye and Sen, Abhijit and Senbanjo, Idowu O. and Sepanlou, Sadaf G. and Shalnova, Svetlana A. and Sharma, Sanjib K. and Shaw, Jonathan E. and Shibuya, Kenji and Shin, Dong Wook and Shin, Youchan and Shiri, Rahman and Siantar, Rosalynn and Sibai, Abla M. and Silva, Antonio M. and Silva, Diego Augusto Santos and Simon, Mary and Simons, Judith and Simons, Leon A. and Sjostrom, Michael and Slowikowska-Hilczer, Jolanta and Slusarczyk, Przemyslaw and Smeeth, Liam and Smith, Margaret C. and Snijder, Marieke B. and So, Hung Kwan and Sobngwi, Eugène and S{\"{o}}derberg, Stefan and Soekatri, Moesijanti Y.E. and Solfrizzi, Vincenzo and Sonestedt, Emily and Song, Yi and S{\o}rensen, Thorkild I.A. and Soric, Maroje and J{\'{e}}rome, Charles Sossa and Soumare, Aicha and Staessen, Jan A. and Starc, Gregor and Stathopoulou, Maria G. and Staub, Kaspar and Stavreski, Bill and Steene-Johannessen, Jostein and Stehle, Peter and Stein, Aryeh D. and Stergiou, George S. and Stessman, Jochanan and Stieber, Jutta and St{\"{o}}ckl, Doris and Stocks, Tanja and Stokwiszewski, Jakub and Stratton, Gareth and Stronks, Karien and Strufaldi, Maria Wany and Sun, Chien An and Sundstr{\"{o}}m, Johan and Sung, Yn Tz and Sunyer, Jordi and Suriyawongpaisal, Paibul and Swinburn, Boyd A. and Sy, Rody G. and Szponar, Lucjan and Tai, E. Shyong and Tammesoo, Mari Liis and Tamosiunas, Abdonas and Tang, Line and Tang, Xun and Tanser, Frank and Tao, Yong and Tarawneh, Mohammed Rasoul and Tarp, Jakob and Tarqui-Mamani, Carolina B. and Taylor, Anne and Tchibindat, Félicité and Theobald, Holger and Thijs, Lutgarde and Thuesen, Betina H. and Tjonneland, Anne and Tolonen, Hanna K. and Tolstrup, Janne S. and Topbas, Murat and Top{\'{o}}r-Madry, Roman and Torrent, Maties and Toselli, Stefania and Traissac, Pierre and Trichopoulou, Antonia and Trichopoulos, Dimitrios and Trinh, Oanh T.H. and Trivedi, Atul and Tshepo, Lechaba and Tulloch-Reid, Marshall K. and Tuomainen, Tomi Pekka and Tuomilehto, Jaakko and Turley, Maria L. and Tynelius, Per and Tzotzas, Themistoklis and Tzourio, Christophe and Ueda, Peter and Ukoli, Flora A.M. and Ulmer, Hanno and Unal, Belgin and Uusitalo, Hannu M.T. and Valdivia, Gonzalo and Vale, Susana and Valvi, Damaskini and van der Schouw, Yvonne T. and Van Herck, Koen and Van Minh, Hoang and van Rossem, Lenie and van Valkengoed, Irene G.M. and Vanderschueren, Dirk and Vanuzzo, Diego and Vatten, Lars and Vega, Tomas and Velasquez-Melendez, Gustavo and Veronesi, Giovanni and Verschuren, WM Monique and Verstraeten, Roosmarijn and Victora, Cesar G. and Viegi, Giovanni and Viet, Lucie and Viikari-Juntura, Eira and Vineis, Paolo and Vioque, Jesus and Virtanen, Jyrki K. and Visvikis-Siest, Sophie and Viswanathan, Bharathi and Vollenweider, Peter and Voutilainen, Sari and Vrdoljak, Ana and Vrijheid, Martine and Wade, Alisha N. and Wagner, Aline and Walton, Janette and Mohamud, Wan Nazaimoon Wan and Wang, Ming Dong and Wang, Qian and Wang, Ya Xing and Wannamethee, S. Goya and Wareham, Nicholas and Weerasekera, Deepa and Whincup, Peter H. and Widhalm, Kurt and Widyahening, Indah S. and Wiecek, Andrzej and Wijga, Alet H. and Wilks, Rainford J. and Willeit, Johann and Wilsgaard, Tom and Wojtyniak, Bogdan and Wong, Jyh Eiin and Wong, Tien Yin and Woo, Jean and Woodward, Mark and Wu, Frederick C. and Wu, Jianfeng and Wu, Shou Ling and Xu, Haiquan and Xu, Liang and Yamborisut, Uruwan and Yan, Weili and Yang, Xiaoguang and Yardim, Nazan and Ye, Xingwang and Yiallouros, Panayiotis K. and Yoshihara, Akihiro and You, Qi Sheng and Younger-Coleman, Novie O. and Yusoff, Ahmad F. and Zainuddin, Ahmad A. and Zambon, Sabina and Zdrojewski, Tomasz and Zeng, Yi and Zhao, Dong and Zhao, Wenhua and Zheng, Yingfeng and Zhou, Maigeng and Zhu, Dan and Zimmermann, Esther and Cisneros, Julio Zuñiga},
    month = {7},
    volume = {5},
    publisher = {eLife Sciences Publications Ltd},
    doi = {10.7554/ELIFE.13410},
    issn = {2050084X},
    pmid = {27458798}
}

@article{Masiero2019APositioning,
    title = {{A comparison of UWB and motion capture uav indoor positioning}},
    year = {2019},
    journal = {International Archives of the Photogrammetry, Remote Sensing and Spatial Information Sciences - ISPRS Archives},
    author = {Masiero, A. and Fissore, F. and Antonello, R. and Cenedese, A. and Vettore, A.},
    number = {2/W13},
    month = {6},
    pages = {1695--1699},
    volume = {42},
    publisher = {International Society for Photogrammetry and Remote Sensing},
    doi = {10.5194/ISPRS-ARCHIVES-XLII-2-W13-1695-2019,},
    issn = {16821750}
}

@article{Plastropoulos2025AHangar,
    title = {{A comprehensive review of robotics-aided aircraft non-destructive inspection towards the smart hangar}},
    year = {2025},
    journal = {The Aeronautical Journal},
    author = {Plastropoulos, A. and Zolotas, A. and Avdelidis, N. P.},
    month = {7},
    pages = {1--32},
    publisher = {Cambridge University Press},
    doi = {10.1017/AER.2025.10048},
    issn = {0001-9240}
}

@article{Schroeer2018AScenarios,
    title = {{A Real-Time UWB Multi-Channel Indoor Positioning System for Industrial Scenarios}},
    year = {2018},
    journal = {IPIN 2018 - 9th International Conference on Indoor Positioning and Indoor Navigation},
    author = {Schroeer, Guido},
    month = {11},
    publisher = {Institute of Electrical and Electronics Engineers Inc.},
    isbn = {9781538656358},
    doi = {10.1109/IPIN.2018.8533792}
}

@article{Merriaux2017APerformance,
    title = {{A Study of Vicon System Positioning Performance}},
    year = {2017},
    journal = {Sensors 2017, Vol. 17, Page 1591},
    author = {Merriaux, Pierre and Dupuis, Yohan and Boutteau, Rémi and Vasseur, Pascal and Savatier, Xavier},
    number = {7},
    month = {7},
    pages = {1591},
    volume = {17},
    publisher = {Multidisciplinary Digital Publishing Institute},
    doi = {10.3390/S17071591},
    issn = {1424-8220},
    pmid = {28686213}
}

@article{Zafari2019ATechnologies,
    title = {{A Survey of Indoor Localization Systems and Technologies}},
    year = {2019},
    journal = {IEEE Communications Surveys and Tutorials},
    author = {Zafari, Faheem and Gkelias, Athanasios and Leung, Kin K.},
    number = {3},
    pages = {2568--2599},
    volume = {21},
    publisher = {Institute of Electrical and Electronics Engineers Inc.},
    doi = {10.1109/COMST.2019.2911558},
    issn = {1553877X},
    arxivId = {1709.01015}
}

@article{Moupfouma2013AircraftDamages,
    title = {{Aircraft Structure Paint Thickness and Lightning Swept Stroke Damages}},
    year = {2013},
    journal = {SAE International Journal of Aerospace},
    author = {Moupfouma, Fidele and Moupfouma, Fidele},
    number = {2},
    month = {9},
    pages = {392--398},
    volume = {6},
    publisher = {SAE International},
    doi = {10.4271/2013-01-2135},
    issn = {19463901}
}

@misc{Airbus2025AutoCADDrawings,
    title = {{AutoCAD 3-view aircraft drawings}},
    year = {2025},
    booktitle = {Airbus},
    author = {{Airbus}},
    url = {https://aircraft.airbus.com/en/customer-care/fleet-wide-care/airport-operations-and-aircraft-characteristics/autocad-3-view-aircraft-drawings}
}

@misc{UbisenseAviationGuide,
    title = {{Aviation {\&} MRO Asset Tracking - The Ultimate Guide}},
    author = {{Ubisense}},
    url = {https://ubisense.com/aviation-mro-asset-tracking/}
}

@misc{Statista2025AviationType,
    title = {{Aviation industry - aircraft fleet by type}},
    year = {2025},
    author = {{Statista}},
    month = {4},
    url = {https://www.statista.com/statistics/573231/aviation-industry-aircraft-fleet-by-type/},
    language = {English}
}

@misc{AviationA320,
    title = {{Aviation Stack Exchange - What is the total wing surface area of the A320?}},
    url = {https://aviation.stackexchange.com/questions/54511/what-is-the-total-wing-surface-area-of-the-a320}
}

@misc{TheBoeingCompany2025CADPurposes,
    title = {{CAD 3-View Drawings for Airport Planning Purposes}},
    year = {2025},
    booktitle = {The Boeing Company},
    author = {{The Boeing Company}},
    url = {https://www.boeing.com/commercial/airports/3-view}
}

@techreport{ForrestCoin-or/Cbc,
    title = {{Coin-or/Cbc}},
    author = {Forrest, John and Ralphs, Ted and Vigerske, Stefan and Santos, Haroldo Gambini and Forrest, John and Hafer, Lou and Kristjansson, Bjarni and {jpfasano} and {EdwinStraver} and {Jan-Willem} and Lubin, Miles and {rlougee} and {a-andre} and {jpgoncal1} and Brito, Samuel and {h-i-gassmann} and {Cristina} and Saltzman, Matthew and {tosttost} and Pitrus, Bruno and MATSUSHIMA, Fumiaki and Vossler, Patrick and SWGY, Ron @ and {to-st}},
    doi = {10.5281/ZENODO.13347261}
}

@techreport{Airbus2024CustomerServices,
    title = {{Customer Services Technical Data Support and Services}},
    year = {2024},
    author = {{Airbus}},
    month = {6},
    url = {https://aircraft.airbus.com/sites/g/files/jlcbta126/files/2025-01/AC_A320_0624.pdf},
    address = {Blagnac Cedex}
}

@article{Meyer2024DesignFilter,
    title = {{Design of a low-cost optical motion capture system using a multi-camera configuration and an asynchronous extended Kalman filter}},
    year = {2024},
    journal = {MATEC Web of Conferences},
    author = {Meyer, Zakariya and Pretorius, Arnold},
    editor = {du Preez, W. and Becker, T. and Modiba, R. and Chauke, H. and Dzogbewu, T. and Mostert, R. and Maringa, M. and Bissett, H. and Koen, W. and ter Haar, G. and van der Merwe, H. and Moema, J. and Botha, N. and Fisher, C. and Wang, H.},
    month = {12},
    pages = {04010},
    volume = {406},
    publisher = {EDP Sciences},
    doi = {10.1051/MATECCONF/202440604010},
    issn = {2261-236X}
}

@article{Park2023FusionRTLS,
    title = {{Fusion localization for indoor airplane inspection using visual inertial odometry and ultrasonic RTLS}},
    year = {2023},
    journal = {Scientific Reports},
    author = {Park, Ingyoon and Cho, Sangook},
    number = {1},
    month = {12},
    pages = {1--13},
    volume = {13},
    publisher = {Nature Research},
    doi = {10.1038/s41598-023-43425-y},
    issn = {20452322},
    pmid = {37872183}
}

@article{Hayduk1973HailSurfaces,
    title = {{Hail Damage to Typical Aircraft Surfaces}},
    year = {1973},
    journal = {Journal of Aircraft},
    author = {Hayduk, Robert J.},
    number = {1},
    month = {1},
    pages = {52--55},
    volume = {10},
    doi = {10.2514/3.60196},
    issn = {00218669}
}

@article{Jong2024HandlingPersonnel,
    title = {{Handling, inspection and repair of aircraft composites: a pilot study on the awareness of maintenance personnel}},
    year = {2024},
    journal = {The Aeronautical Journal},
    author = {Jong, C. M. and Comer, A. J. and Chatzi, A. V. and Kourousis, K. I.},
    number = {1320},
    month = {2},
    pages = {211--229},
    volume = {128},
    publisher = {Cambridge University Press},
    doi = {10.1017/AER.2023.28},
    issn = {0001-9240}
}

@article{Fatima2022HighSafety,
    title = {{High Precision Indoor Positioning System for Mobile Asset Management and Safety}},
    year = {2022},
    journal = {Engineering Proceedings 2022, Vol. 20, Page 37},
    author = {Fatima, Hafiza Sundus and Zaman, Laeeq Uz and Zia, Huma and Khurram, Muhammad},
    number = {1},
    month = {8},
    pages = {37},
    volume = {20},
    publisher = {Multidisciplinary Digital Publishing Institute},
    doi = {10.3390/ENGPROC2022020037},
    issn = {2673-4591}
}

@book{Panero1992HumanStandards,
    title = {{Human dimensions and interior space: A source book of design reference standards}},
    year = {1992},
    author = {Panero, Julius and Zelnik, Martin},
    month = {3},
    pages = {71--74},
    publisher = {Whitney Library of Design},
    isbn = {978-0823072712}
}

@misc{ClearpathRoboticsHuskySpecifications,
    title = {{Husky A300 Robot Specifications}},
    author = {{Clearpath Robotics}},
    url = {https://clearpathrobotics.com/husky-a300-unmanned-ground-vehicle-robot/}
}

@article{Gerwen2022IndoorFusion,
    title = {{Indoor Drone Positioning: Accuracy and Cost Trade-Off for Sensor Fusion}},
    year = {2022},
    journal = {IEEE Transactions on Vehicular Technology},
    author = {Gerwen, Jono Vanhie Van and Geebelen, Kurt and Wan, Jia and Joseph, Wout and Hoebeke, Jeroen and De Poorter, Eli},
    number = {1},
    month = {1},
    pages = {961--974},
    volume = {71},
    publisher = {Institute of Electrical and Electronics Engineers Inc.},
    doi = {10.1109/TVT.2021.3129917},
    issn = {19399359}
}

@article{Moenck2025IndustryOpportunities,
    title = {{Industry 5.0 in aircraft production and MRO: challenges and opportunities}},
    year = {2025},
    journal = {CEAS Aeronautical Journal},
    author = {Moenck, Keno and Koch, Julian and Rath, Jan Erik and B{\"{u}}sch, Lukas and Gierecker, Johann and K{\"{a}}hler, Falko and Kalscheuer, Florian and Masuhr, Christian and Kipping, Johann and Pr{\"{u}}nte, Philipp and Schoepflin, Daniel and Eschen, Henrik and Wulff, Lukas Antonio and Rodeck, Rebecca and Wende, Gerko and Gomse, Martin and Sch{\"{u}}ppstuhl, Thorsten},
    month = {4},
    pages = {1--25},
    publisher = {Springer},
    doi = {10.1007/s13272-025-00832-3},
    issn = {18695590}
}

@article{Rojas2015InnovativeCracks,
    title = {{Innovative NDT Technique Based on Ferrofluids for Detection of Surface Cracks}},
    year = {2015},
    journal = {Journal of Nondestructive Evaluation},
    author = {Rojas, J. I. and Cabrera, B. and Musterni, G. and Nicolas, J. and Tristancho, J. and Crespo, D.},
    number = {4},
    month = {11},
    pages = {1--12},
    volume = {34},
    publisher = {Springer New York LLC},
    doi = {10.1007/s10921-015-0309-5},
    issn = {15734862}
}

@article{Chen2014InspectionDamage,
    title = {{Inspection intervals optimization for aircraft composite structures considering dent damage}},
    year = {2014},
    journal = {Journal of Aircraft},
    author = {Chen, Xi and Ren, He and Bil, Cees},
    number = {1},
    month = {2},
    pages = {303--309},
    volume = {51},
    publisher = {American Institute of Aeronautics and Astronautics Inc.},
    doi = {10.2514/1.C032377},
    issn = {15333868}
}

@article{Leugner2018LessonsApplication,
    title = {{Lessons learned: Indoor Ultra-Wideband localization systems for an industrial IoT application}},
    year = {2018},
    journal = {Proceedings of the 3rd KuVS/GI Expert Talk on Localization},
    author = {Leugner, Swen and Hellbr{\"{u}}ck, Horst},
    pages = {2--4},
    doi = {10.24355/DBBS.084-201807191346-0}
}

@article{Pugliese2021LiDAR-AidedFuselages,
    title = {{LiDAR-Aided Relative and Absolute Localization for Automated UAV-based Inspection of Aircraft Fuselages}},
    year = {2021},
    journal = {IEEE International Conference on Multisensor Fusion and Integration for Intelligent Systems},
    author = {Pugliese, Roland and Konrad, Thomas and Abel, Dirk},
    publisher = {Institute of Electrical and Electronics Engineers Inc.},
    doi = {10.1109/MFI52462.2021.9591172}
}

@techreport{Fisher1982Lightning80,
    title = {{Lightning Attachemnt Patterns and Flight Conditions for Storm Hazards ' 80}},
    year = {1982},
    author = {Fisher, Bruce and Keyser, Gerald and Deal, Perry},
    url = {https://ntrs.nasa.gov/api/citations/19830005806/downloads/19830005806.pdf},
    institution = {National Aeronautics and Space Administration}
}

@misc{TransmonEngineeringLtdManagingSpeeds,
    title = {{Managing Forklift Speeds}},
    author = {{Transmon Engineering Ltd}},
    url = {https://www.transmon.co.uk/news/managing-forklift-speed-limits/}
}

@article{Lvov2023MobileOn-The-Go,
    title = {{Mobile MoCap: Retroreflector Localization On-The-Go}},
    year = {2023},
    journal = {IEEE International Conference on Automation Science and Engineering},
    author = {Lvov, Gary and Zolotas, Mark and Hanson, Nathaniel and Allison, Austin and Hubbard, Xavier and Carvajal, Michael and Padir, Taskin},
    volume = {2023-August},
    publisher = {IEEE Computer Society},
    isbn = {9798350320695},
    doi = {10.1109/CASE56687.2023.10260562},
    issn = {21618089}
}

@article{Wang2024Multipath-AssistedLearning,
    title = {{Multipath-Assisted Single-Anchor Localization via Deep Variational Learning}},
    year = {2024},
    journal = {IEEE Transactions on Wireless Communications},
    author = {Wang, Tianyu and Li, Yuxiao and Liu, Junchen and Hu, Keke and Shen, Yuan},
    number = {8},
    pages = {9113--9128},
    volume = {23},
    publisher = {Institute of Electrical and Electronics Engineers Inc.},
    doi = {10.1109/TWC.2024.3359047},
    issn = {15582248}
}

@misc{SpanTechNarrowHangars,
    title = {{Narrow Body Aircraft Hangars}},
    booktitle = {airport-suppliers.com},
    author = {{SpanTech}},
    url = {https://www.airport-suppliers.com/product/narrow-body-aircraft-hangars/}
}

@article{Wallis1966OnFacilities,
    title = {{On the Aerodynamics of “Hangar” Type Engine Test Facilities}},
    year = {1966},
    journal = {Journal of the Royal Aeronautical Society},
    author = {Wallis, R. A. and Ruglen, N.},
    number = {662},
    month = {2},
    pages = {312--320},
    volume = {70},
    publisher = {Cambridge University Press (CUP)},
    doi = {10.1017/S0368393100082183},
    issn = {0368-3931}
}

@article{Rahimian2017OptimalSystems,
    title = {{Optimal Camera Placement for Motion Capture Systems}},
    year = {2017},
    journal = {IEEE Transactions on Visualization and Computer Graphics},
    author = {Rahimian, Pooya and Kearney, Joseph K.},
    number = {3},
    month = {3},
    pages = {1209--1221},
    volume = {23},
    publisher = {IEEE Computer Society},
    doi = {10.1109/TVCG.2016.2637334},
    issn = {10772626},
    pmid = {27959813}
}

@article{Murtagh2021OutdoorMeta-analysis,
    title = {{Outdoor Walking Speeds of Apparently Healthy Adults: A Systematic Review and Meta-analysis}},
    year = {2021},
    journal = {Sports medicine (Auckland, N.Z.)},
    author = {Murtagh, Elaine M. and Mair, Jacqueline L. and Aguiar, Elroy and Tudor-Locke, Catrine and Murphy, Marie H.},
    number = {1},
    month = {1},
    pages = {125--141},
    volume = {51},
    publisher = {Sports Med},
    doi = {10.1007/S40279-020-01351-3},
    issn = {1179-2035},
    pmid = {33030707}
}

@misc{HusarionPantherSpecifications,
    title = {{Panther Robot Specifications}},
    author = {{Husarion}},
    url = {https://husarion.com/manuals/panther/overview/}
}

@article{Karadeniz2020PreciseNodes,
    title = {{Precise UWB-based localization for aircraft sensor nodes}},
    year = {2020},
    journal = {AIAA/IEEE Digital Avionics Systems Conference - Proceedings},
    author = {Karadeniz, Cansu Gozde and Geyer, Fabien and Multerer, Thomas and Schupke, Dominic},
    month = {10},
    volume = {2020-October},
    publisher = {Institute of Electrical and Electronics Engineers Inc.},
    isbn = {9781728198255},
    doi = {10.1109/DASC50938.2020.9256793},
    issn = {21557209}
}

@article{Mitchell2011PuLPPython,
    title = {{PuLP : A Linear Programming Toolkit for Python}},
    year = {2011},
    author = {Mitchell, S. and O’Sullivan, Michael and Dunning, Iain},
    url = {https://optimization-online.org/wp-content/uploads/2011/09/3178.pdf}
}

@article{Tatale2018Quadcopter:Testing,
    title = {{Quadcopter: design, construction and testing}},
    year = {2018},
    journal = {Tatale, Omkar, et al. "Quadcopter: design, construction and testing." International Journal for Research in Engineering Application {\&} Management},
    author = {Tatale, Omkar and Anekar, Nitinkumar and Phatak, Supriya and Sarkale, Suraj},
    number = {Special Issue AMET-2018},
    pages = {1--7},
    doi = {10.18231/2454-9150.2018.1386}
}

@article{Bahnemann2019RevisitingProblem,
    title = {{Revisiting Boustrophedon Coverage Path Planning as a Generalized Traveling Salesman Problem}},
    year = {2019},
    journal = {Springer Proceedings in Advanced Robotics},
    author = {B{\"{a}}hnemann, Rik and Lawrance, Nicholas and Chung, Jen Jen and Pantic, Michael and Siegwart, Roland and Nieto, Juan},
    month = {7},
    pages = {277--290},
    volume = {16},
    publisher = {Springer Science and Business Media B.V.},
    doi = {10.1007/978-981-15-9460-1{\_}20},
    arxivId = {1907.09224v1}
}

@misc{BostonDynamicsSpotSpecifications,
    title = {{Spot Rorot Specifications}},
    author = {{Boston Dynamics}},
    url = {https://support.bostondynamics.com/s/article/Spot-Specifications-49916}
}

@article{Baaran2009StudyStructures,
    title = {{Study on visual inspection of composite structures}},
    year = {2009},
    journal = {European Aviation Safety Agency},
    author = {Baaran, Jens},
    month = {7},
    url = {https://www.easa.europa.eu/sites/default/files/dfu/EASA_REP_RESEA_2007_3.pdf}
}

@article{Plastropoulos2024TheAviation,
    title = {{The ‘hangar of the future’ for sustainable aviation}},
    year = {2024},
    journal = {The Aeronautical Journal},
    author = {Plastropoulos, A. and Fan, I. S. and Avdelidis, N. P. and Angus, J. P. and Maggiore, J. and Atkinson, H.},
    number = {1329},
    pages = {2429--2450},
    volume = {128},
    publisher = {Cambridge University Press},
    doi = {10.1017/AER.2024.79},
    issn = {0001-9240}
}

@article{Hansen2022UAVSolution,
    title = {{UAV Trajectory Evaluation in Large Industrial Environments: A Cost-Effective Solution}},
    year = {2022},
    journal = {2022 European Control Conference, ECC 2022},
    author = {Hansen, Jakob Grimm and Hei{\ss}, Micha and Koz{\l}owski, Michał and Kayacan, Erdal},
    pages = {1336--1341},
    publisher = {Institute of Electrical and Electronics Engineers Inc.},
    isbn = {9783907144077},
    doi = {10.23919/ECC55457.2022.9838352}
}

@misc{UbisenseUbisenseSolution,
    title = {{Ubisense and MRO Drone launch ‘Smart Hangar’ solution}},
    author = {{Ubisense}},
    url = {https://ubisense.com/ubisense-and-mro-drone-launch-worlds-first-smart-hangar-solution/}
}

@misc{ForkliftsafetyWhatOperations,
    title = {{What Are The Recommended Speed Regulations For Forklift Operations?}},
    author = {{Forklift safety}},
    url = {https://www.forkliftsafety.com/2023/09/16/what-are-the-recommended-speed-regulations-for-forklift-operations/}
}

\end{document}